\documentclass[journal]{IEEEtran}
\usepackage{amsmath,amsfonts}
\usepackage{algorithmic}
\usepackage{algorithm}
\usepackage{array}
\usepackage[caption=false,font=normalsize,labelfont=sf,textfont=sf]{subfig}
\usepackage{textcomp}
\usepackage{stfloats}
\usepackage{url}
\usepackage{verbatim}
\usepackage{graphicx}
\usepackage{cite}
\usepackage{xcolor}
\usepackage{hyperref}
\usepackage{multirow}
\usepackage{makecell}
\usepackage{nicematrix}
\usepackage[table,dvipsnames,x11names]{xcolor}
\usepackage{float}
\newcommand{\ieno}{\textit{i.e.}}
\newcommand{\egno}{\textit{e.g.}}
\usepackage{siunitx}
\usepackage{pifont}
\newcommand{\cmark}{\ding{51}}
\newcommand{\xmark}{\ding{55}}
\usepackage{amssymb}
\usepackage{threeparttable}
\usepackage{booktabs, tabularx}
\usepackage[table]{xcolor}
\usepackage[utf8]{inputenc}
\usepackage[T1]{fontenc}
\usepackage{newunicodechar}

\begin{document}

\title{UDPNet: Unleashing Depth-based Priors for Robust Image Dehazing}

\author{
Zengyuan Zuo,~
Junjun~Jiang,~\IEEEmembership{Senior Member,~IEEE},~
Gang~Wu,~
and Xianming Liu,~\IEEEmembership{Member,~IEEE}
\thanks{
Z. Zuo, J. Jiang, G. Wu, and X. Liu are with the School of Computer Science and Technology,
Harbin Institute of Technology, Harbin 150001, China.
E-mail: \{24s103286@stu.hit.edu.cn, jiangjunjun@hit.edu.cn, gwu@hit.edu.cn, csxm@hit.edu.cn\}.
Corresponding author: Junjun Jiang.
}
}

\markboth{Journal of \LaTeX\ Class Files,~Vol.~14, No.~8, August~2021}%
{Zuo \MakeLowercase{\textit{et al.}}: UDPNet: Unleashing Depth-based Priors for Robust Image Dehazing}

\maketitle

\begin{abstract}
Image dehazing has witnessed significant advancements with the development of deep learning models. However, most existing methods focus solely on single-modal RGB features, neglecting the inherent correlation between scene depth and haze distribution. Even those that jointly optimize depth estimation and image dehazing often suffer from suboptimal performance due to inadequate utilization of accurate depth information. In this paper, we present UDPNet, a general framework that leverages depth-based priors from a large-scale pretrained depth estimation model DepthAnything V2 to boost existing image dehazing models. Specifically, our architecture comprises two key components: the Depth-Guided Attention Module (DGAM) adaptively modulates features via lightweight depth-guided channel attention, and the Depth Prior Fusion Module (DPFM) enables hierarchical fusion of multi-scale depth map features by dual sliding-window multi-head cross-attention mechanism. These modules ensure both computational efficiency and effective integration of depth priors. Moreover, the depth priors empower the network to dynamically adapt to varying haze densities, illumination conditions, and domain gaps across synthetic and real-world data. Extensive experimental results demonstrate the effectiveness of our UDPNet, outperforming the state-of-the-art methods on popular dehazing datasets, with PSNR improvements of 0.85 dB on SOTS-indoor, 1.19 dB on Haze4K, and 1.79 dB on NHR. Our proposed solution establishes a new benchmark for depth-aware dehazing across various scenarios. Pretrained models and codes are released at our project ~\url{https://github.com/Harbinzzy/UDPNet}.
\end{abstract}


\section{Introduction}

\begin{figure}[tp]
    \centering
    \includegraphics[width=1\linewidth]{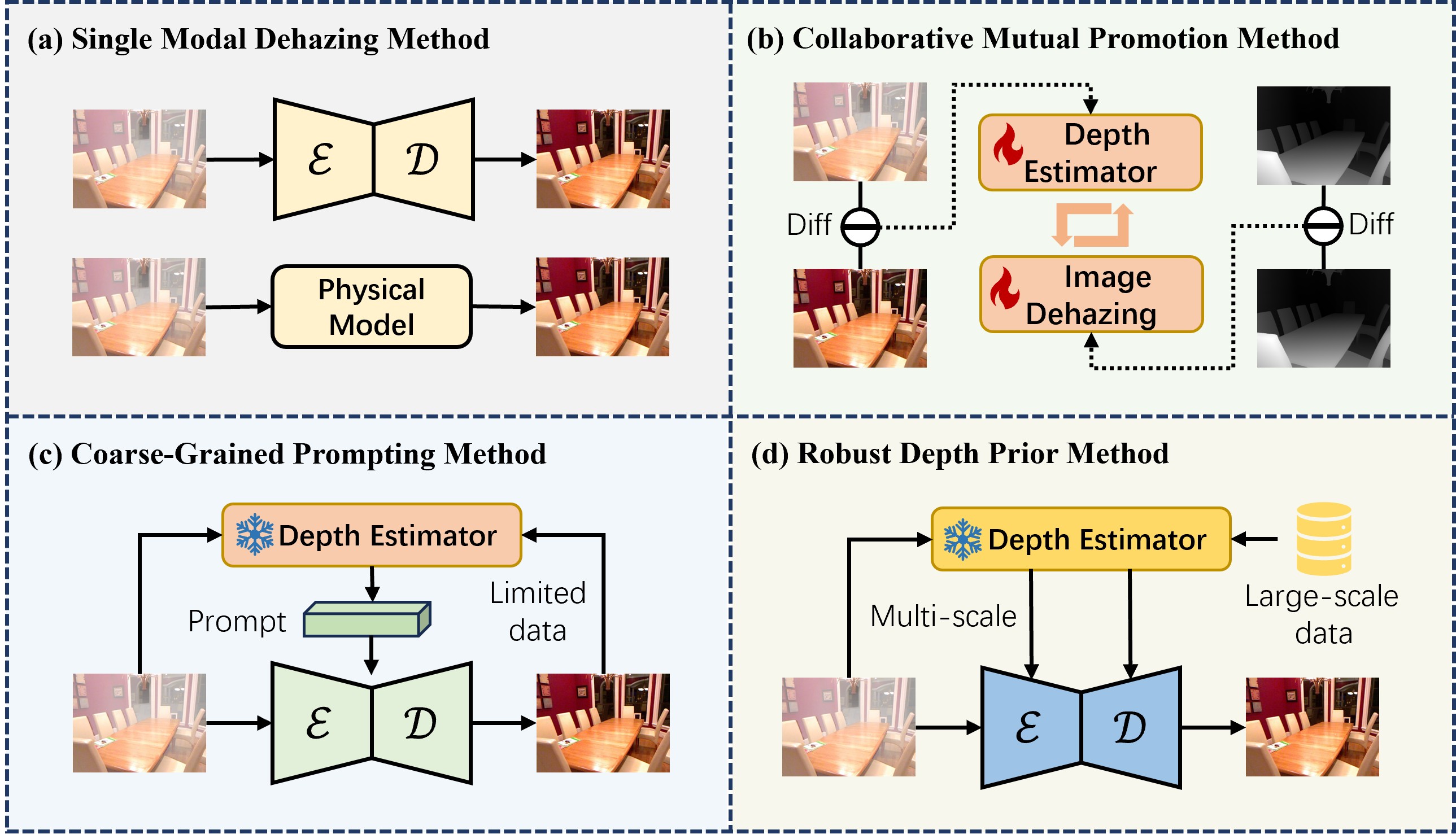}
    \caption{\textbf{Illustration of Representative Image Dehazing Frameworks.} \textbf{(a)} Approaches based on deep learning~\cite{dong2020multi, wu2021contrastive, cui2024revitalizing, qiu2023mb, jin2025mb} or physical models~\cite{he2010single, zhu2015fast, he2012guided, ju2019idgcp}. \textbf{(b)} The collaborative mutual promotion between depth estimation and image dehazing~\cite{zhang2024depth}. \textbf{(c)} The depth-consistent prompt dehazing network~\cite{wang2024selfpromer}. \textbf{(d)} Our UDPNet utilizes the robust and reliable depth estimation model trained on large-scale datasets and effectively integrates it into a multi-scale hierarchical network, achieving excellent generalization.}
    \label{fig:first}
\end{figure}

\IEEEPARstart{I}{mage} dehazing, a fundamental low-level image restoration task, plays an essential preprocessing role in enhancing the reliability of downstream high-level vision systems (\egno, object detection, semantic segmentation). According to the Atmospheric Scattering Model (ASM)~\cite{middleton1957vision}, the degradation process of hazy images can be mathematically described as: 
\begin{equation}
H^c(x, y) = J^c(x, y) \cdot t(x, y) + A^c \cdot \left(1 - t(x, y)\right)
\end{equation}
where $(x, y)$ represents the pixel coordinate, $c \in \{r, g, b\}$ denotes the color channel, $H$ denotes the hazy image, $J$ denotes the corresponding clear image, $A$ is the atmospheric light. The transmission map is usually modeled as $t(x, y) = e^{-\beta \cdot d(x, y)}$ with the scattering coefficient $\beta$ and the scene depth $d(x, y)$. 
It follows from the above equation that the estimation accuracy of the transmission map and atmospheric light is crucial to traditional dehazing methods~\cite{zhu2015fast, he2012guided, ju2019idgcp, he2010single}, which rely on reasonable assumptions or statistical priors (see Fig.~\ref{fig:first} (a)). However, the complexity of real-world environments often leads to violations of the underlying priors, thereby reducing the robustness of traditional dehazing methods. For instance, He \emph{et al.}~\cite{he2010single} assumed that at least one color channel has very low intensity in most haze-free patches, which may not hold in bright regions such as the sky or objects with high reflectance, resulting in color distortions. Similarly, priors based on color-line distributions~\cite{zhu2015fast} or contrast assumptions~\cite{ju2019idgcp} can break down in cases where the scene contains complex textures, non-uniform illumination, or artificial light sources. Moreover, the estimation of atmospheric light can be unreliable in scenes with strong sunlight, multiple scattering, or dense haze, further amplifying the reconstruction error. These limitations highlight the inherent fragility of prior-based dehazing when applied to real-world conditions.

With the advent of deep learning, various variations of Convolutional Neural Networks (CNNs)~\cite{dong2020multi, wu2021contrastive, cui2023image, cui2024revitalizing} and Transformers~\cite{liu2021swin, zamir2022restormer, guo2022image, qiu2023mb, jin2025mb} have emerged for image dehazing by learning the strong statistical regularities. They typically compute a sequence of features from the single-modal RGB inputs and directly reconstruct the clear ones (see Fig.~\ref{fig:first} (a)). Such approaches have achieved state-of-the-art results on benchmark datasets by leveraging advanced architectural designs such as multi-scale information fusion~\cite{liu2019griddehazenet, cui2023image, cui2024revitalizing}, multi-stage pipelines, sophisticated variants of convolution~\cite{liu2020trident, luo2023lkd, cui2024omni}, attention mechanisms~\cite{liu2019griddehazenet, cui2024revitalizing}. Nevertheless, dehazing remains an ill‑posed inverse problem in which feature estimation errors can markedly degrade restoration quality. Since RGB observations alone provide incomplete and often ambiguous cues under heavy haze, relying exclusively on them limits the model’s ability to resolve uncertainty and correct erroneous predictions. Inspired by the human visual system’s capacity to integrate and reconcile multisensory information for robust perception, we argue that effective dehazing requires the joint exploitation of complementary modalities (\egno, depth maps~\cite{zhang2024depth}, near-infrared~\cite{li2025hierarchical} or semantic maps~\cite{zhang2021semantic}) to enhance robustness and generalization in haze removal. The ASM further provides a theoretical basis, as haze density is intrinsically correlated with scene depth, suggesting a natural synergy between dehazing and depth estimation. Leveraging depth-based priors can thus supply structural guidance and holistic awareness of haze distribution. Following this insight, several recent works~\cite{wang2024selfpromer, zhang2024depth} integrated depth into dehazing frameworks. For instance, DCMPNet~\cite{zhang2024depth} (see Fig.~\ref{fig:first} (b)) incorporated the depth estimation and image dehazing model within a unified framework in a dual-task-driven manner, while exhibiting reduced effectiveness in outdoor and real-world scenarios. SelfPromer~\cite{wang2024selfpromer} (see Fig.~\ref{fig:first} (c)) extracted the depth difference features between hazy and clear images as prompts to improve haze removal. Although the perception quality has been enhanced, the distortion metric is poor.

\begin{figure}
    \centering
    \includegraphics[width=1\linewidth]{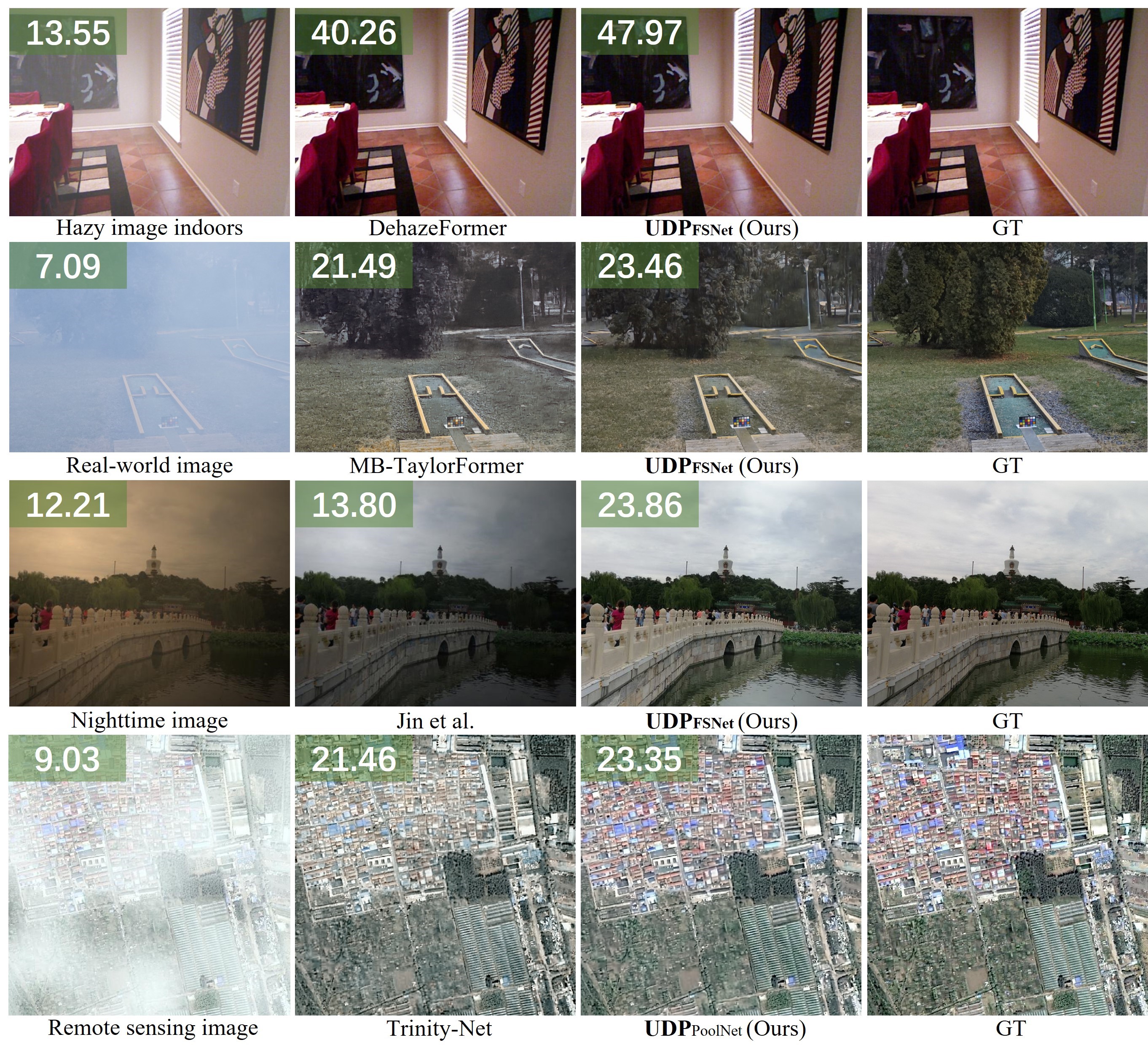}
    \caption{Comparisons of dehazing methods across varying conditions. Our framework demonstrates consistent performance improvements compared to representative methods.}
    \label{fig:second}
\end{figure}

However, there still exist three main challenges in practical image dehazing: (1) \textit{The acquired depth information is usually inaccurate.}  Prior studies have shown that larger datasets and more powerful models provide richer cues beyond basic labeling. Yet existing depth-based methods, whether relying on pretrained depth estimation models~\cite{birkl2023midas} with limited training data or lightweight models trained from scratch~\cite{yang2022depth, zhang2024depth}, still struggle to cope with complex hazy scenes such as nonhomogeneous, dense, or nighttime haze. (2) \textit{The potential of depth information has not been fully exploited.} Although the difference self-prompt approach~\cite{wang2024selfpromer} improved perceptual quality, it still performed poorly on distortion-oriented metrics. Similarly, directly integrating depth features at the tail of a UNet backbone~\cite{zhang2024depth} failed to yield satisfactory results. More recently, PromptHaze~\cite{ye2025prompthaze} adopted a relatively straightforward design by injecting depth prompt features as an additional input, which limited its ability to capture the complex interactions between depth and haze. (3) \textit{Consistently high-performing dehazing across both daytime and nighttime remains challenging.} Real-world nighttime haze is further complicated by multiple scattering, uneven illumination, glow, blur, and hidden noise, which make it more challenging than daytime haze. Existing state-of-the-art daytime dehazing methods~\cite{liu2021swin, zamir2022restormer, guo2022image, qiu2023mb} and nighttime counterparts~\cite{jin2023enhancing, liu2023nighthazeformer} are not explicitly designed to handle both daytime and nighttime scenarios. Thus, developing a universal model that can jointly address daytime haze, nighttime haze, and diverse haze conditions is still an urgent challenge, as illustrated in Fig.~\ref{fig:second}.

To address these challenges, we turn to the fundamental limitations posed by depth estimation and its integration with visual features in the context of image dehazing. Depth estimation, while a promising cue, is often hindered by inherent inaccuracies and limited generalization across complex hazy scenes (Issue 1). In response, we employ DepthAnything V2~\cite{depthanything_v2}, a state-of-the-art pretrained model, which offers robust and generalizable depth cues. To fully exploit the potential of depth as a structural prior (Issue 2), we propose two lightweight plug-and-play modules that effectively integrate depth with visual representations. The Depth-Guided Attention Module (DGAM) adaptively modulates feature maps via depth-informed channel attention, ensuring efficient interaction between depth and image content. Complementing this, the Depth Prior Fusion Module (DPFM) enables hierarchical fusion of multi-scale depth features with visual representations, facilitating richer structural guidance and enhancing robustness in dehazing tasks. This dual-module approach bridges the gap between depth and image representations, providing a unified solution for diverse hazy scenarios.

In recent years, foundation models have profoundly influenced various image restoration tasks. However, to the best of our knowledge, no existing work has attempted to jointly address dehazing, nighttime dehazing, and other image restoration tasks using DepthAnything V2, a model inherently related to haze distribution and global structure priors. This work is motivated by the observation that depth-based priors can enhance robustness and reconstruction quality across various scenarios.
 In summary, our main contributions are as follows:
\begin{enumerate}
    \item[$\bullet$] We introduce the large-scale pretrained depth estimation model for image dehazing. The systematical incorporation of priors enables our network to generalize to a wide range of hazy image scenarios.
    \item[$\bullet$] We propose a general framework that integrates the depth maps into the multi-scale image dehazing networks. Meanwhile, we propose a depth-based prior fusion strategy that balances effectiveness and computational efficiency.  
    \item[$\bullet$] Our UDPNet model achieve state-of-the-art performance on various dehazing datasets, exhibiting consistent improvements across both daytime/nighttime conditions and synthetic/real-world scenarios, while also demonstrating potential in other image restoration tasks.
\end{enumerate}

The remainder of this paper is organized as follows. Sec.~\ref{sec:Related} reviews existing image dehazing methods, with a particular focus on depth-based approaches. Sec.~\ref{sec:Analysis} presents our observations and motivation, along with preliminary experiments. Sec.~\ref{sec:methods} details the proposed UDPNet, and Sec.~\ref{sec:exp} presents experimental results across popular datasets along with discussions on limitations. Finally, Sec.~\ref{sec:conclusion} concludes the paper.

\section{Related Work}
\label{sec:Related}
\subsection{Image Dehazing}
Image dehazing methods are typically divided into prior-based, learning-based, and hybrid approaches. Because of its ill-posed nature, prior-based methods rely on the atmospheric scattering model and handcrafted priors, such as the dark channel prior (DCP)~\cite{he2010single}, color attenuation prior (CAP)~\cite{zhu2015fast}, and BCCR~\cite{meng2013efficient}, to estimate the transmission map and atmospheric light. While effective under ideal conditions that meet their assumptions, these approaches struggle to generalize in complex haze scenarios. In contrast, learning-based methods leverage data-driven architectures such as CNNs and Transformers to achieve superior dehazing performance. Some early models focus on learning transmission maps~\cite{ren2016single, cai2016dehazenet} or jointly learning transmission maps and atmospheric light~\cite{pang2018visual, zhang2018densely}. In parallel, Dehaze-RetinexGAN~\cite{wang2025dehaze} embed the Retinex-based decomposition model into the dehazing network to improve generalization ability. 

However, these methods are susceptible to cumulative errors arising from inaccurate estimations. To avoid this, more recent works~\cite{liu2019griddehazenet, wang2021ensemble, chen2024dea, cui2024omni} incorporate hierarchical multi-scale representations, attention mechanisms, large-kernel convolutions, enabling effective dehazing without reliance on the physical model. GridDehazeNet~\cite{liu2019griddehazenet} employs a three-stage attention-based multi-scale framework to achieve end-to-end dehazing. EPDN~\cite{qu2019enhanced}, a pioneering framework embedding GAN and the enhancer module, generates perceptually pleasing images with realistic colors and fine details. FFA-Net~\cite{qin2020ffa} introduces the feature attention (FA) and attention-based feature fusion (FFA) to effectively integrate information across different levels. Cui \emph{et al.}~\cite{cui2023image, cui2024revitalizing, su2025prior} design the hierarchical multi-input multi-output pure convolutional network to recover image sharpness and textures. C\textsuperscript{2}PNet~\cite{zheng2023curricular} and PHATNet~\cite{tsai2025phatnet} incorporate physics priors into the feature space, enhancing interpretability by aligning feature representations with the hazing process. Dehamer~\cite{guo2022image}, DehazeFormer~\cite{song2023vision} and MB-TaylorFormer~\cite{qiu2023mb} leverage Transformers to model long-range dependencies. Concurrently, with the advancement of data-driven methods in dehazing performance, network complexity also increases. To mitigate the reliance on large-scale training data, hybrid approaches~\cite{mo2022dca} were proposed, integrating inherent image priors with the representation power of deep neural networks. For example, PGH$^2$Net~\cite{su2025prior} introduced triple priors (\ieno, bright channel prior, dark channel prior and histogram equalization prior) to UNet-like architecture. Methods like ~\cite{fang2025guided} utilize transformation of color space (YCbCr) to guide the RGB space. Furthermore, SelfPromer~\cite{wang2024selfpromer} and PromptHaze~\cite{ye2025prompthaze} explore the fruitful combination of depth-based priors with deep dehazing network.

In addition to these architectural advances, several complementary directions have emerged, including methods for nighttime dehazing~\cite{yan2020nighttime, jin2023enhancing}, haze image generation approaches~\cite{wang2025learning, wu2023ridcp} for real-world scenarios, and recent efforts toward universal image restoration~\cite{jiang2024survey}. For example, NightHazeFormer~\cite{liu2023nighthazeformer} integrates dual priors into the transformer decoder, effectively guiding the network to learn abundant prior
features. NightHaze~\cite{lin2025nighthaze} introduces a self-prior learning paradigm, leveraging MAE-like representations to improve nighttime dehazing performance. Wang \emph{et al.}~\cite{wang2025learning} propose HazeGen and DiffDehaze, a diffusion-based hazing–dehazing pipeline. Together, these efforts broaden the scope of dehazing research and lay a solid foundation for addressing more challenging restoration tasks. 

\subsection{Depth Priors for Image Dehazing}
While end-to-end dehazing networks relying solely on RGB features have advanced significantly, they often struggle with ambiguous scene structures due to insufficient depth constraints. In contrast, approaches incorporating depth maps~\cite{yang2022depth, yang2022self, ye2025prompthaze}  provide distinct advantages by leveraging embedded structural priors.
The intrinsic physical correlation further highlights the potential. Lee \emph{et al.}~\cite{lee2020cnn} propose a CNN-based simultaneous dehazing and depth estimation network with a shared encoder and multiple decoders. Similarly, Yang \emph{et al.}~\cite{yang2022depth} and Cheng \emph{et al.}~\cite{cheng2021two} propose the depth-aware methods to fuse the depth features to the dehazing network. Due to the absence of accurate depth maps during inference and the possible residual haze remaining, Wang \emph{et al.}~\cite{wang2024selfpromer} propose a continuous self-prompting inference strategy, which iteratively refines the dehazing results toward clearer outputs. Furthermore, PromptHaze~\cite{ye2025prompthaze} introduces a dehazing strength factor to achieve a controllable step-by-step dehazing process, thereby alleviating the impact of cumulative errors. Recently, Zhang \emph{et al.}~\cite{zhang2024depth} integrate lightweight depth estimation model and image dehazing to form a dual-task collaborative optimization. Some novel methods~\cite{yang2022self, wu2023ridcp, liu2025peie, ye2025prompthaze} perform online synthesis of realistic hazy images based on depth maps and physical models, facilitating supervised training for real-world dehazing. These innovations inspire us to develop an effective and general dehazing framework that leverages depth priors as guidance. Moreover, our work proposes a general depth map fusion scheme, which is validated through experiments on previous algorithms~\cite{cui2023image, cui2024revitalizing, cui2025exploring, potlapalli2023promptir, cui2025adair}. In the following sections, we will introduce the detailed design of our methods.

\subsection{DepthAnything vs. Other Depth Estimation Models}
DepthAnything~\cite{depthanything} and its successor DepthAnything V2~\cite{depthanything_v2} represent a significant shift in monocular depth estimation through a precise synthetic and large‑scale pseudo‑labeled paradigm. Unlike MiDaSv3.1~\cite{birkl2023midas} or AdaBins~\cite{bhat2021adabins}, which are exclusively trained on manually annotated datasets (\egno, NYU‑D~\cite{silberman2012indoor}, KITTI~\cite{geiger2013vision}), DepthAnything first distilled a high-capacity DINOv2-Giant~\cite{oquab2023dinov2} teacher model on five high‐fidelity synthetic datasets, and then generated pseudo‐depth for 62 million in‑the‑wild real images, effectively bridging the domain gap between synthetic and real-world data distributions. This methodology leads to superior performance in handling thin structures and transparent objects, and significantly improves generalization across complex and challenging scenes. The global depth scene information, coupled with the refined sharpness of depth discontinuities, effectively serves as an auxiliary to image dehazing tasks. Our framework ensures a seamless integration of depth-based priors with dehazing networks.

\begin{figure}[t]
    \centering
    \includegraphics[width=\linewidth]{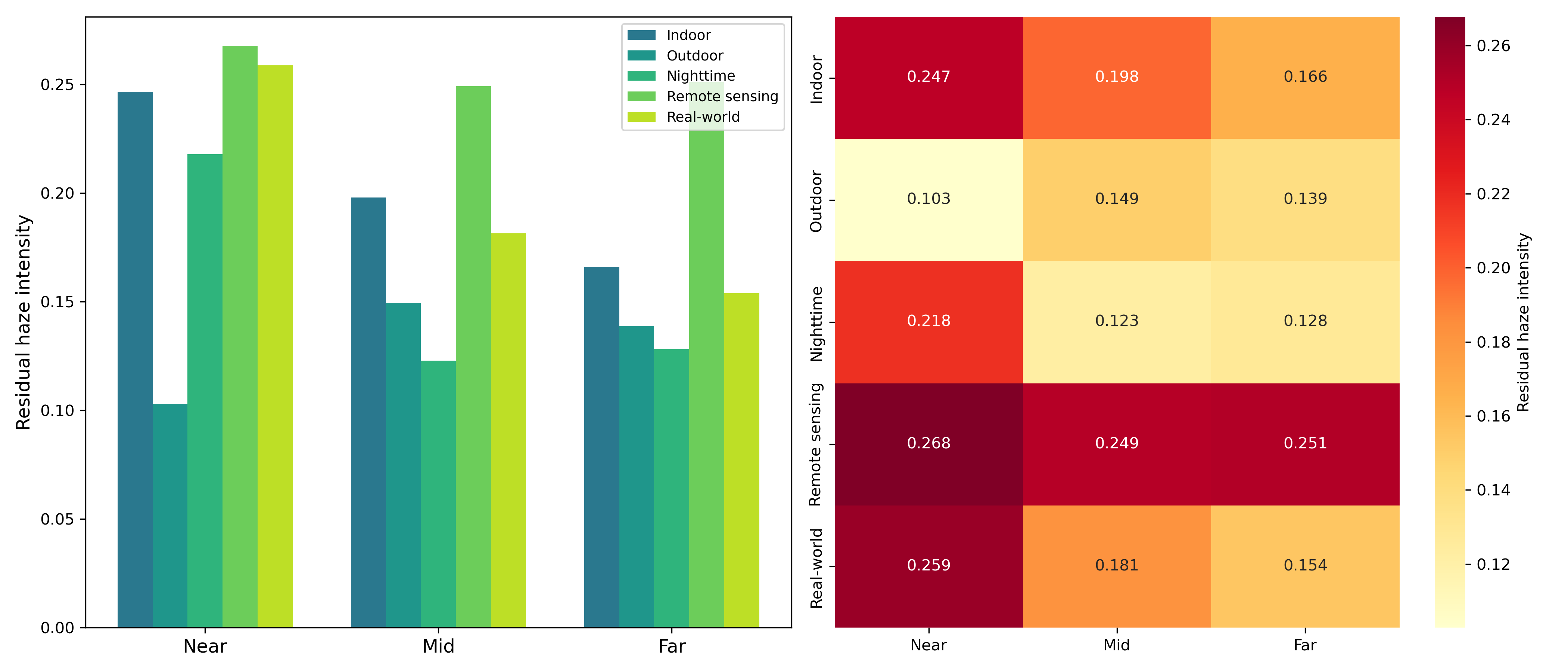}
    \caption{Cross-dataset statistical visualization of the residual haze (i.e., haze density) across depth ranges highlighting scene-dependent variations and depth–haze correlations in indoor, outdoor, nighttime, remote sensing, and real-world scenarios.}
    \label{fig:show}
\end{figure}

\begin{figure}[t]
    \vspace{-3mm}
    \centering
    \includegraphics[width=\linewidth]{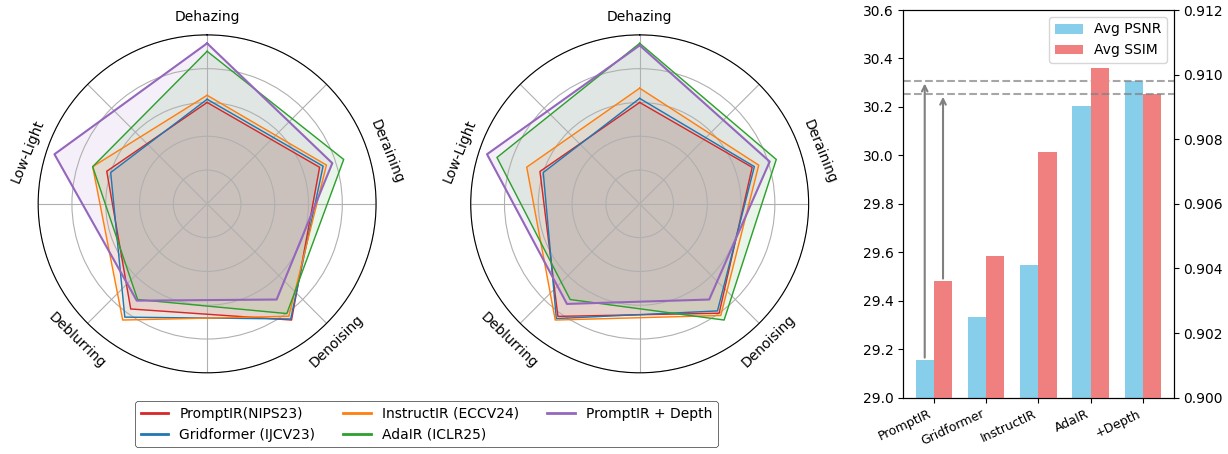}
    \caption{Comparative analysis of the impact of depth map assistance on various image restoration tasks. Our experiments show that incorporating depth maps yields substantial gains in dehazing and low-light enhancement, demonstrating their potential for both daytime and nighttime dehazing. Moreover, the overall performance achieves state-of-the-art results. Depth maps are generated using DepthAnything-small~\cite{depthanything}, and our dehazing network, PromptIR~\cite{potlapalli2023promptir}, takes both RGB and depth inputs rather than relying solely on RGB features.}
    \label{fig:third}
\end{figure}

\section{Preliminary Analysis}
\label{sec:Analysis}
In this paper, we propose that the depth prior, often overlooked, is powerful global information inherent to image dehazing. Mathematically, the atmospheric scattering model links the transmission map to depth via
\begin{equation}
     t(x,y) = e^{-\beta d(x,y)}.
\end{equation}
This implies that deeper regions should appear more heavily degraded. However, in practice this monotonic trend does not always hold across different scenes. To better understand this discrepancy, we analyze the relationship between depth and residual haze, where residual haze is defined as the absolute difference between the hazy input and its corresponding clear image. Specifically, for each dataset we compute a residual haze map, partition the normalized depth into three equal bins (Near, Mid, Far), and calculate the average residual haze intensity within each bin. Fig.~\ref{fig:show} presents a cross-dataset statistical analysis. Interestingly, the trends vary significantly across scene categories. Outdoor datasets exhibit stronger haze density in distant regions, which is consistent with physical scattering intuition, whereas indoor datasets show stronger residuals at near distances. Nighttime datasets present different residuals, reflecting the compounded degradations from haze and illumination. Remote sensing scenes reveal consistently strong haze density across all depths, while real-world datasets demonstrate a mixed trend. These observations highlight that the relationship between depth and haze density is scene-dependent, emphasizing the need to incorporate depth priors for adaptive dehazing, rather than relying on handcrafted priors or rigid physical formula-based constraints.

Meanwhile, we further argue that depth estimation models have the capacity to jointly handle dehazing and low-light enhancement, and this capacity can be further enhanced through elaborate design. In all-in-one image restoration, a single network is expected to address diverse degradations and adaptively decide how to restore the input. Depth information provides valuable guidance, particularly for dehazing and low-light scenarios, by separating foreground from background and enabling more accurate restoration across varying depths. This improves robustness in complex real-world conditions. To validate the effectiveness of our proposed method, we perform preliminary experiments for verification. We experiment with a dedicated all-in-one image restoration network~\cite{potlapalli2023promptir} evaluating on the five-degradation scenarios~\cite{jiang2024survey} (modifying the original RGB input by concatenating it with depth maps extracted by DepthAnything-small~\cite{depthanything}). The results are presented in Fig.~\ref{fig:third}. It can be seen that the average PSNR achieves state-of-the-art performance, with particularly significant improvements on the dehazing and low-light enhancement tasks. This indicates that the depth maps extracted by DepthAnything inherently contribute to both daytime and nighttime image dehazing. Additionally, the depth-based approach shows significant potential for enhancing performance in other image restoration tasks, deserving further exploration.

\begin{figure*}[t]
    \centering
    \includegraphics[width=1\linewidth]{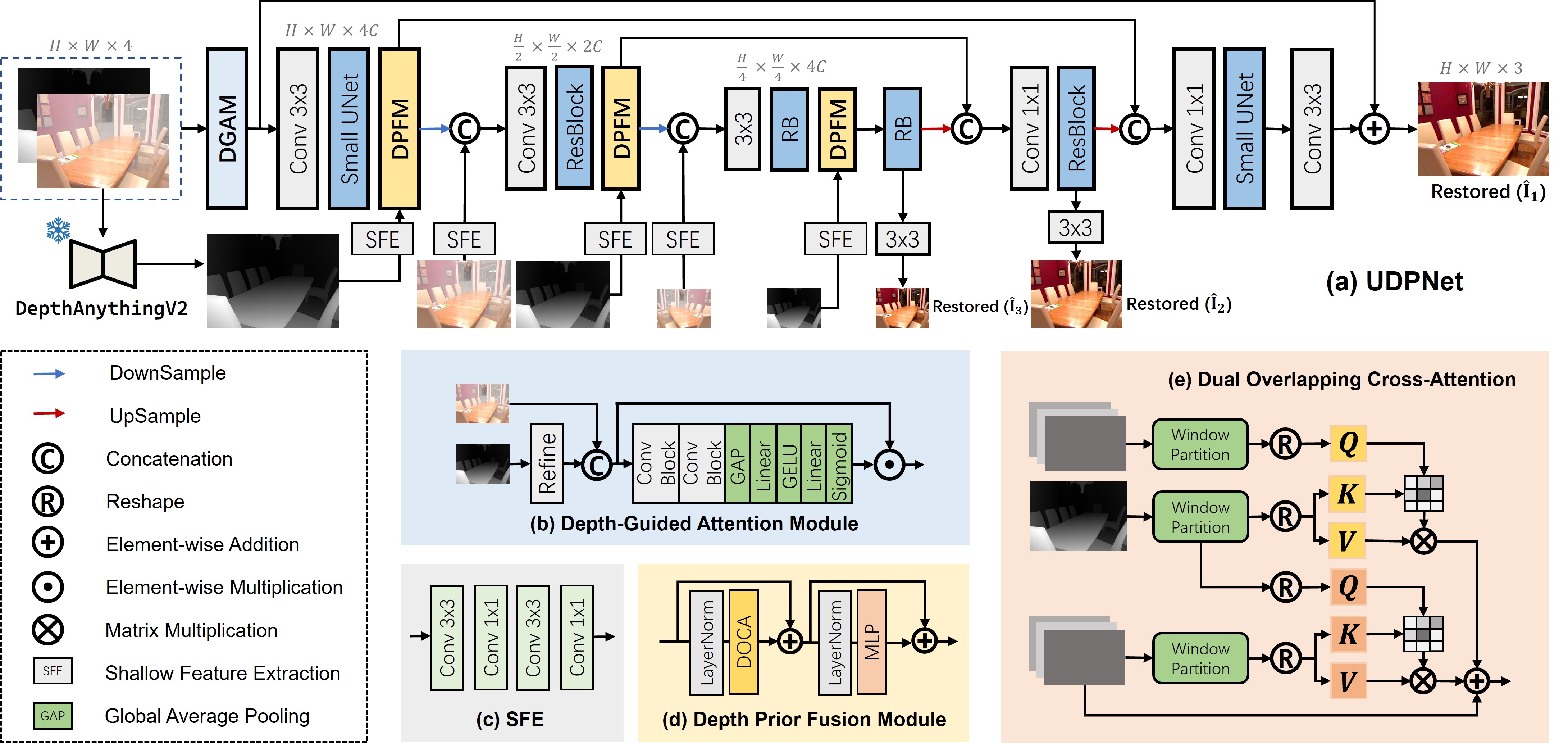}
    \caption{The architecture of our proposed UDPNet (a) is inspired by the symmetric hierarchical encoder-decoder design~\cite{cui2023image}. UDPNet is a specific instantiation of our UDP (Unleashing Depth-based Priors) paradigm, which is realized by integrated sub-modules, \ieno, DGAM (b) and DPFM (d). DGAM focuses on channel-level fusion between depth-based priors and the input image. DPFM incorporates a multi-scale, multi-head cross-attention mechanism with overlapping windows and dual attention pathways at the spatial level.}
    \label{fig:fourth}
\end{figure*}

\section{Methodology}
\label{sec:methods}

In this section, we introduce our proposed depth-based framework, as illustrated in Fig.~\ref{fig:fourth}. We first present an overall pipeline of our UDPNet. We then delineate details of our designs: the Depth-Guided Attention Module (detailed in Sec.~\ref{sec:dgam}) and the Depth Prior Fusion Module (detailed in Sec.~\ref{sec:dpfm}). The loss functions are introduced in the final part.

\subsection{Overall Architecture}
In image dehazing, depth information has proven to be a valuable cue for recovering structural details and improving robustness, especially in complex hazy environments. However, traditional methods often struggle to fully exploit depth priors due to challenges in accurately estimating depth and integrating it effectively with image features. Motivated by this, we aim to integrate depth-based priors into a U-shaped backbone for haze removal. Our approach leverages a pretrained depth estimation model, DepthAnything V2, to generate robust depth maps, which are then used to guide the feature extraction process, enhancing the network's ability to capture geometric and structural information. By incorporating these priors into a multi-scale architecture, we ensure that both high-level and low-level features are effectively utilized.

Specifically, the network is established by incorporating depth-based priors with a U-shaped CNN-based backbone for haze removal. Both the encoder and decoder networks comprise three scales, efficiently learning hierarchical representations. Specifically, given any degraded image $\mathbf{I} \in \mathbb{R}^{3 \times H \times W}$, UDPNet first leverages a frozen DepthAnything V2 to generate robust depth maps with the dimensions of $1 \times H \times W$, where $H \times W$ represents spatial locations. The input image and corresponding depth map are processed by the Depth-Guided Attention Module (DGAM), which adapts the features based on depth information. The resulting features are then passed through an asymmetric encoder-decoder structure. 

Each stage of the encoder includes a Depth Prior Fusion Module (DPFM), designed to better integrate depth priors with the visual features. To ensure effective integration, this module is applied in the encoder, as depth map information is most beneficial for capturing geometric cues and structural details during the early stages of feature extraction. Introducing the module in the decoder would inject low-frequency information from the depth map into the high-frequency image restoration process, potentially hindering fine-grained detail restoration. Next, the depth-enhanced features are passed through a three-scale decoder, progressively restoring high-resolution features. During training, a multi-output strategy is employed, where low-resolution clean images are predicted using $3 \times 3$ convolutions and skip connections. For brevity, Fig.~\ref{fig:fourth}(a) only shows the top-level image skip connection. Although our approach can be applied to various dehazing baselines, we demonstrate the framework on FSNet~\cite{cui2023image} for clarity.

\subsection{Depth Priors Fusion}
Effective image dehazing requires not only removal of haze but also faithful recovery of underlying scene structures, spatial consistency, and depth-aware contrasts. While RGB images provide rich color information, they often lack sufficient cues to resolve structural ambiguities, especially in complex haze conditions. Depth priors, capturing geometric layouts and relative distances within a scene, offer complementary information that can guide the network toward more accurate restoration. However, existing methods either rely on simple concatenation of depth maps with image features or apply simple integration strategies, which fail to fully exploit the rich structural knowledge embedded in depth cues. Motivated by this, we aim to systematically incorporate depth priors from a large-scale pretrained model into the dehazing pipeline, enhancing both fidelity and generalization across diverse haze.

Depth priors provide essential cues for recovering scene structures, depth-aware contrasts, and spatial consistency in degraded images. As a large-scale foundation model trained on diverse image distributions, DepthAnything V2 offers rich geometric knowledge to guide restoration. Leveraging its understanding of spatial layouts and depth relationships, our method incorporates depth-based priors to enhance fidelity and generalizability. We design two modules to integrate these depth cues into the conventional dehazing pipeline.

\subsubsection{Depth-Guided Attention Module}
\label{sec:dgam}
Given a hazy input image $\mathbf{I} \in \mathbb{R}^{3 \times H \times W}$ concatenated with its estimated depth map $\mathbf{D} \in \mathbb{R}^{1 \times H \times W}$, we first refine the depth map using a refinement module to suppress noise and enhance geometric cues. The refinement is implemented as a three-layer convolutional network with ReLU activations:
\begin{equation}
\mathbf{\hat{D}} = \text{Refinement}(\mathbf{D}),
\end{equation}
where $\mathbf{\hat{D}}$ denotes the optimized depth map.

The refined depth map is concatenated with the RGB image and fed into the subsequent modules to extract depth-aware features. To further enhance feature discrimination, a depth-guided attention mechanism is applied. We first employ a $3 \times 3$ convolution followed by instance normalization and GELU activation to process the concatenated RGB-D input and produce intermediate feature representations. Subsequent modules capture global context through adaptive average pooling, generate channel-wise attention weights via two fully connected layers with a GELU nonlinearity, and apply a sigmoid activation. The resulting attention weights are then used to reweight the input features, thereby emphasizing informative channels and suppressing less relevant ones. Formally, the process can be expressed as
\begin{equation}
    \mathbf{F}_{\text{DGAM}} = [\mathbf{I}, \mathbf{\hat{D}}] \odot \sigma(\text{FC}_2(\text{GELU}(\text{FC}_1(\text{GAP}(\mathbf{F}_{\text{conv}}))))),
\end{equation}
where $[\cdot, \cdot]$ is a concatenation operator, $\odot$ denotes element-wise multiplication, $F_{\text{conv}}$ denotes output features of the convolution block, $\text{GAP}$ is global average pooling, $\text{FC}_1$ and $\text{FC}_2$ are fully connected layers, and $\sigma$ is the sigmoid activation. By integrating the refined depth map and RGB image through this DGAM, as depicted in Fig.~\ref{fig:fourth}(b), the network obtains depth-aware shallow features that benefit subsequent restoration stages. The objective experimental results validate that the proposed DGAM is more effective than a simple concatenation strategy. As detailed in Section \ref{sec:Ablation_Study}, employing DGAM yields a performance gain on the Haze4K dataset compared to naive feature stacking, demonstrating the advantage of adaptive, depth-guided channel attention.

\subsubsection{Depth Prior Fusion Module}
\label{sec:dpfm}
The representations encoded in shallow and deep features differ substantially, as they correspond to entirely different receptive fields. This disparity makes it essential to unleash depth priors across multiple feature scales. Inspired by SwinIR~\cite{liang2021swinir} and Restormer~\cite{zamir2022restormer}, we propose a depth prior fusion module (DPFM) to inject estimated depth priors into the multi-scale restoration backbone via a dual cross-attention mechanism defined on overlapping local windows, as shown in Fig.~\ref{fig:fourth}(d). Given an intermediate feature map $\mathbf{X}\in\mathbb{R}^{H\times W\times C}$ and a single-channel depth prior $\mathbf{D}\in\mathbb{R}^{H\times W\times 1}$, the depth map is first projected by shallow feature extract to produce depth map features $\mathbf{F}_D\in\mathbb{R}^{H\times W\times C}$. Both $\mathbf{X}$ and $\mathbf{F}_D$ are partitioned into local windows of size $M\times M$ with overlap ratio $r$ (so the effective overlap-window side length is $M_{\mathrm{ov}}=(1+r)\times M$). Zero-padding is first applied to make $H$ and $W$ divisible by $M$. Queries are then partitioned into non-overlapping windows of size $M\times M$, while keys and values are unfolded from larger overlapping windows of size $M_{\mathrm{ov}}\times M_{\mathrm{ov}}$. This overlap enables each query window to incorporate neighboring context at a controllable cost. As a result, it propagates geometric cues across adjacent regions while maintaining computational tractability.

Within each local window, we instantiate two complementary cross-attention branches. The \emph{depth-guided} branch where depth features act as queries and image features provide keys and values, which emphasizes geometric constraints and guides the restoration with structural priors. The \emph{feature-guided} branch where image features act as queries and depth features provide keys and values, which enhances the consistency of fine image details with depth information. Together, these two branches form an interaction mechanism that jointly exploits depth and image cues. For a window $w$ (vectorized to $\mathbb{R}^{M^2\times C}$), let $h$ be the number of heads and $d=C/h$ the per-head dimension; denote by $B$ a learnable relative-position bias. The two branches are written as
\begin{equation}
\text{Attn}_{D \rightarrow X} = 
\text{Softmax}\!\left(\frac{Q_D K_X^\top}{\sqrt{d}} + B\right)V_X,
\end{equation}
\begin{equation}
\text{Attn}_{X \rightarrow D} = 
\text{Softmax}\!\left(\frac{Q_X K_D^\top}{\sqrt{d}} + B\right)V_D,
\end{equation}
with
\[
Q_D=\mathbf{F}_{D,w}W_Q^D,\quad K_X=\mathbf{X}_w W_K^X,\quad V_X=\mathbf{X}_w W_V^X,
\]
and analogously $Q_X,K_D,V_D$ for the other branch. The window output aggregates both branches with a residual connection:
\begin{equation}
\mathbf{Y}_w = \text{Attn}_{D \rightarrow X} + \text{Attn}_{X \rightarrow D} + \mathbf{X}_w.
\end{equation}

All window outputs are merged to the full spatial resolution and passed through a pointwise feed-forward network (FFN) with LayerNorm and residual connections to produce the fused feature $\mathbf{Y}\in\mathbb{R}^{H\times W\times C}$, which can be formulated as
\begin{equation}
    \mathbf{Y} = \text{DOCA}(\text{LN}(\mathbf{X})) + \mathbf{X}, \quad
    \mathbf{Y} = \text{MLP}(\text{LN}(\mathbf{Y})) + \mathbf{Y}.
\end{equation}

In practice, a learnable relative-position bias is introduced to encode spatial priors. Compared to global cross-attention with quadratic cost $\mathcal{O}((HW)^2)$, 
the window overlapping design yields complexity $\mathcal{O}(HW \cdot M \cdot M_{\mathrm{ov}})$, since each $M^2$ query interacts with $M_{\mathrm{ov}}^2$ key/value tokens. Although slightly higher than standard non-overlapping window cross attention ($\mathcal{O}(HW \cdot M^2)$), the complexity remains linear in the number of pixels $HW$ and quadratic only in the local window sizes, thus available for high-resolution image restoration. Furthermore, we insert DPFM blocks at multiple encoder stages, allowing depth-aware priors to be progressively injected into the backbone. This hierarchical design enhances geometric consistency and structural fidelity in the restored images, fusing the
multi-scale depth information at negligible extra computational cost. The ablation results in Section \ref{sec:Ablation_Study} indicate that inserting the DPFM within the encoder stages is more beneficial than placing it in the decoder. This is attributed to the fact that depth priors provide strong structural and geometric cues that are most effective when integrated during the early stages of feature extraction, providing better guidance for subsequent hierarchical processing. Furthermore, the spatial cross-attention mechanism in DPFM proves particularly effective, contributing significantly to the overall state-of-the-art performance achieved on Haze4K.

\subsection{Loss Functions}
Following Cui \emph{et al.}~\cite{cui2023image, cui2024revitalizing}, we employ a multi-scale dual-domain loss to train UDPNet, where the spatial loss is defined as 
$\mathcal{L}_{\text{spatial}}=\sum_{i=1}^{3}\frac{1}{P_i}\|\hat{\mathbf{I}}_i-\mathbf{Y}_i\|_1$
and the frequency-domain loss is 
$\mathcal{L}_{\text{frequency}}=\sum_{i=1}^{3}\frac{1}{S_i}\|[\mathcal{R}(\hat{\mathbf{I}}_i),\mathcal{I}(\hat{\mathbf{I}}_i)]-[\mathcal{R}(\mathbf{Y}_i),\mathcal{I}(\mathbf{Y}_i)]\|_1$, where $i$ denotes the index corresponding to the multiple output branches. $\hat{\mathbf{I}}$ represents the restored image, while $\mathbf{Y}$ denotes the ground truth. The $P$ and $S$ correspond to the total elements for normalization.
The notation $[\cdot, \cdot]$ is a concatenation operator, and $\mathcal{R}$ and $\mathcal{I}$ represent the real and imaginary parts obtained through the fast Fourier transform.

\subsection{Remarks}
Many existing depth-aware dehazing methods incorporate depth cues either by direct concatenation with inputs or through complex schemes such as joint training with depth estimators or prompt-based conditioning. While effective in some cases, these strategies do not always ensure a stable and practical integration of depth information into image features.

In contrast, our framework systematically integrates depth priors through two plug-and-play modules. DGAM modulates feature representations adaptively based on depth-driven channel attention, thereby ensuring that structural cues are enhanced, while DPFM performs multi-scale fusion to promote spatial consistency across different representation levels. Together, these modules provide a structured way to inject depth information into the restoration pipeline.

Experimental results confirm that our approach yields consistent improvements over existing methods, particularly in various haze conditions where purely RGB-based models may lack geometric cues. Moreover, the fusion is implemented with residual connections and attention-based modulation, allowing the network to down-weight unreliable depth features and fall back to RGB information when depth cues are inconsistent with evidence. This design ensures graceful improvements rather than catastrophic failure when depth estimation is imperfect, as shown in Table~\ref{tab:depth_map}. Finally, while this work focuses on dehazing, the proposed depth-aware fusion design is potentially extendable to advancing universal image restoration.

\section{Experiments}
\label{sec:exp}
In this section, we conduct extensive experiments to evaluate the effectiveness and generalization of our proposed UDPNet. We first introduce the datasets, metrics, and implementation details. Then we perform quantitative and qualitative assessments. In the tables, the baseline results are marked in blue for clarity. The best and second-best results are highlighted in boldface and underlined, respectively.

\subsection{Datasets and Metrics}
\label{sec:datasets}

\begin{table}[t]
\centering
\caption{Summary of datasets used for training and evaluation.}
\label{tab:datasets}
\resizebox{1\columnwidth}{!}{
\begin{tabular}{l|l|l|l}
\hline
\textbf{Dataset} & \textbf{Type} & \textbf{Scale / Characteristics} & \textbf{Usage} \\
\hline
\multicolumn{4}{c}{\textbf{Single image dehazing settings}} \\
\hline
RESIDE~\cite{li2018benchmarking} & Synthetic & ITS (indoor), OTS (outdoor), SOTS & Indoor/Outdoor \\
Haze4K~\cite{liu2021synthetic} & Synthetic & 4000 paired images, more realistic & Indoor/Outdoor \\
NH-HAZE~\cite{ancuti2020nh} & Real-world & 55 paired images, non-homogeneous & Real-world \\
Dense-Haze~\cite{ancuti2019dense} & Real-world & 55 paired images, dense & Real-world \\
NHR~\cite{zhang2020nighttime} & Synthetic & 16,146 train pairs, 1,794 test pairs & Nighttime \\
GTA5~\cite{yan2020nighttime} & Synthetic & Large-scale, game-engine rendered & Nighttime \\
SateHaze1k~\cite{huang2020single} & Synthetic & 1,200 hazy/clear pairs, satellite imagery & Remote sensing \\
\hline
\multicolumn{4}{c}{\textbf{All-in-one image restoration settings}} \\
\hline
RESIDE~\cite{li2018benchmarking} & Synthetic & OTS \& SOTS-outdoor & Dehazing \\
Rain100L~\cite{yang2019joint} & Synthetic & 100 training \& 100 testing images & Deraining \\
BSD400~\cite{martin2001database} & Real-world & 400 images from Berkeley segmentation & Denoising \\
WED~\cite{ma2016waterloo} & Real-world & 4,744 high-quality natural images & Denoising \\
GoPro~\cite{nah2017deep} & Real-world & 3,214 pairs from GoPro videos & Deblurring \\
LOL-v1~\cite{wei2018deep} & Real-world & 500 low/normal-light pairs & Low-light \\
\hline
\end{tabular}
}
\end{table}

\textbf{Datasets.} To comprehensively validate the capabilities of the proposed UDPNet, we conduct an evaluation across diverse dehazing benchmarks, as presented in Table~\ref{tab:datasets}.

\textbf{Evaluation Metrics.} Across all datasets, we report PSNR and SSIM~\cite{wang2004image} to quantitatively assess restoration quality. For consistency and comparability, we calculate metrics based on identical implementation employed in prior studies~\cite{cui2024revitalizing, cui2025adair}. On the real-world datasets~\cite{ancuti2019dense,ancuti2020nh}, we further evaluate perceptual similarity using LPIPS~\cite{zhang2018unreasonable}.

\subsection{Implementation Details}
\label{sec:detail}
Our framework is trained on NVIDIA RTX 3090 GPUs. Following the previous methods~\cite{cui2023image}, each patch is randomly flipped horizontally for data augmentation. The initial learning rate is gradually reduced to 1e-6 with cosine annealing~\cite{loshchilov2016sgdr}. Adam ($\beta_1$ = 0.9, $\beta_2$ = 0.999) is used for training. We adopt the FSNet~\cite{cui2023image}, ConvIR~\cite{cui2024revitalizing}, and PoolNet~\cite{cui2025exploring} as our baselines. We follow their original training configurations to ensure a fair comparison. Due to the high resolution and patch size used in paired real-world datasets, we report a resource-aware configuration, where depth-guided attention module yields the visible gains under limited training budget. We also evaluate the UDP paradigm in all-in-one settings (\ieno, five-task configurations~\cite{jiang2024survey}), thereby demonstrating the robustness. We adopt the L1 loss as the sole optimization objective.

\begin{figure*}[t]
    \centering
    \includegraphics[width=1\linewidth]{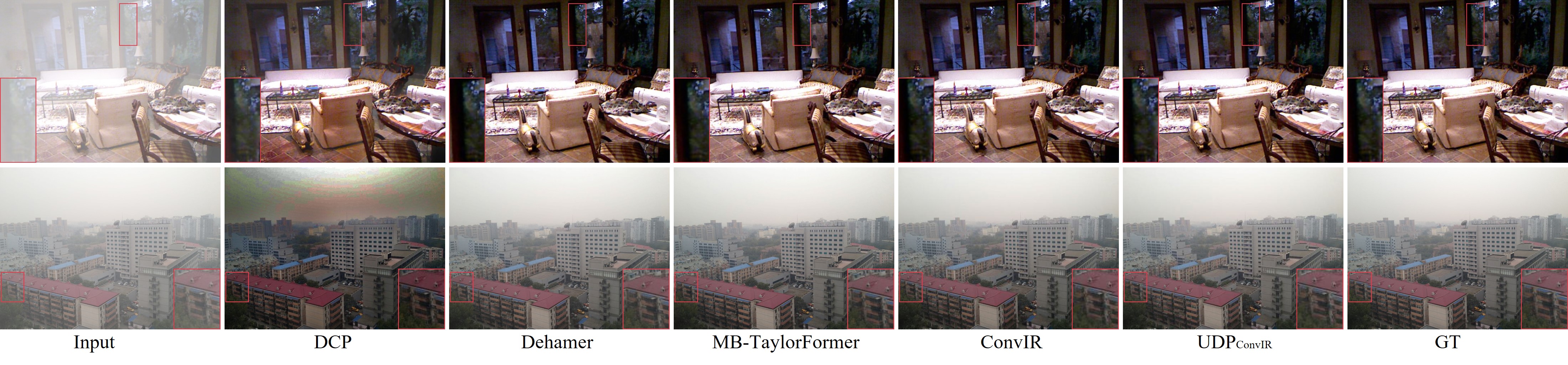}
    \caption{Image dehazing comparisons on the SOTS-indoor~\cite{li2018benchmarking} and SOTS-outdoor datasets~\cite{li2018benchmarking}. Please zoom in for a better view.}
    \label{fig:iots}
\end{figure*}

\begin{table}[t]
\centering
\caption{Quantitative comparison on the SOTS-Indoor and SOTS-Outdoor datasets~\cite{li2018benchmarking}.}
\resizebox{1\columnwidth}{!}{
\begin{tabular}{l|l|cc|cc}
\toprule
\multirow{2}{*}{Method} & \multirow{2}{*}{Venue} 
& \multicolumn{2}{c|}{SOTS-Indoor} 
& \multicolumn{2}{c}{SOTS-Outdoor} \\
\cline{3-6}
&  & PSNR↑ & SSIM↑ & PSNR↑ & SSIM↑ \\
\midrule
DCP~\cite{he2010single}                        & TPAMI'10 & 16.61 & 0.855 & 19.14 & 0.861
\\
FFA-Net~\cite{qin2020ffa}                      & AAAI'20  & 36.39 & 0.989 & 33.57 & 0.984 \\
AECR-Net~\cite{wu2021contrastive}              & CVPR'21  & 37.17 & 0.990 & -     & -     \\
DeHamer~\cite{guo2022image}                    & CVPR'22  & 36.63 & 0.988 & 35.18 & 0.986 \\
DehazeFormer-L~\cite{song2023vision}           & TIP'23   & 40.05 & 0.996 & -     & -     \\
\textcolor{blue}{FSNet}~\cite{cui2023image}    & \textcolor{blue}{TPAMI'23} & \textcolor{blue}{42.45} & \textcolor{blue}{0.997} & \textcolor{blue}{\underline{40.40}} & \textcolor{blue}{\underline{0.997}} \\
MB-TaylorFormer-L~\cite{qiu2023mb}             & ICCV'23  & 42.64 & 0.994 & 38.09 & 0.991 \\
FocalNet~\cite{cui2023focal}                   & ICCV'23  & 40.82 & 0.996 & 37.71 & 0.995 \\
C\textsuperscript{2}PNet~\cite{zheng2023curricular}&CVPR'23&42.56 & 0.995 & 36.68 & 0.990 \\
DEA-Net-CR~\cite{chen2024dea}                  & TIP'24   & 41.31 & 0.995 & 36.59 & 0.990 \\
DCMPNet~\cite{zhang2024depth}                  & CVPR'24 & 42.18 & 0.997 & 36.56 & 0.993\\
GridFormer~\cite{wang2024gridformer}           & IJCV'24 & 42.34 & 0.994 & -     & -   \\
\textcolor{blue}{ConvIR-B}~\cite{cui2024revitalizing}& \textcolor{blue}{TPAMI'24} & \textcolor{blue}{42.72} & \textcolor{blue}{0.997} & \textcolor{blue}{39.42} & \textcolor{blue}{0.996} \\
SFMN~\cite{shen2025spatial}                    & TIP'25   & 41.44 & 0.995 & 37.72 & 0.991 \\
PoolNet-B~\cite{cui2025exploring}              & TIP'25   & 42.01 & 0.997 &  -    &  -    \\
PGH$^2$Net~\cite{su2025prior}                  & AAAI'25  & 41.70 & 0.996 & 37.52 & 0.989 \\
MB-TaylorFormerV2-L~\cite{jin2025mb}           & TPAMI'25 & 42.84 & 0.995 & 39.25 & 0.992 \\
\midrule
\textbf{ConvIR + UDP} & \textbf{Ours} & \underline{43.12} & \underline{0.997} &  40.32& 0.996 \\
\textbf{FSNet + UDP}  & \textbf{Ours} & \textbf{43.30} & \textbf{0.997} & \textbf{40.53} & \textbf{0.997} \\
\bottomrule
\end{tabular}
}
\label{tab:dehaze_all}
\end{table}

\begin{table}[t]
\centering
\caption{Quantitative results on the Haze4K dataset~\cite{liu2021synthetic}.}
\begin{tabular}{l|cc}
\toprule
Method & PSNR↑ & SSIM↑ \\
\midrule
DehazeNet~\cite{cai2016dehazenet}                & 19.12 & 0.84 \\
GridDehazeNet~\cite{liu2019griddehazenet}        & 23.29 & 0.93 \\
FFA-Net~\cite{qin2020ffa}                        & 26.96 & 0.95 \\
DMT-Net~\cite{liu2021synthetic}                  & 28.53 & 0.96 \\
PMNet~\cite{ye2022perceiving}                    & 33.49 & 0.98 \\
MB-TaylorFormer-L~\cite{qiu2023mb}                 & 34.47 & 0.99  \\
\textcolor{blue}{FSNet}~\cite{cui2023image}            & \textcolor{blue}{34.12} & \textcolor{blue}{0.99} \\
GridFormer~\cite{wang2024gridformer}            & 33.27 & 0.99 \\
\textcolor{blue}{ConvIR-B}~\cite{cui2024revitalizing}    & \textcolor{blue}{34.15} & \textcolor{blue}{0.99} \\
DEA-Net-CR~\cite{chen2024dea}                    & 34.25 & 0.99\\
MB-TaylorFormerV2-B~\cite{jin2025mb}             & \underline{34.92} & \underline{0.99} \\
\midrule
\textbf{ConvIR + UDP (Ours)}                    & 34.82 & 0.99 \\
\textbf{FSNet + UDP (Ours)}                     & \textbf{35.31} & \textbf{0.99} \\
\bottomrule
\end{tabular}
\label{tab:haze4k}
\end{table}

\begin{table*}[t]
\centering
\caption{Quantitative results on nighttime image dehazing datasets: GTA5\cite{yan2020nighttime} and NHR\cite{zhang2020nighttime}}
\resizebox{\textwidth}{!}{
\begin{tabular}{l|cccccccccc}
\toprule
\multicolumn{11}{c}{\textbf{The GTA5 dataset}~\cite{yan2020nighttime}} \\
\hline
&GS$^\dagger$ \cite{li2015nighttime}
&MRP$^\dagger$ \cite{zhang2017fast}
&Ancuti \textit{et al}$^\dagger$ \cite{ancuti2016night}
&Yan \textit{et al}$^\dagger$ \cite{yan2020nighttime}
&CycleGAN \cite{zhu2017unpaired}
&Jin \textit{et al}$^\dagger$ \cite{jin2023enhancing}
&\textcolor{blue}{ConvIR-B} \cite{cui2024revitalizing}
&\textcolor{blue}{PoolNet-B} \cite{cui2025exploring}
&\textbf{PoolNet + UDP}
&\textbf{ConvIR + UDP} \\
\midrule
PSNR↑ & 21.02 & 20.92 & 20.59 & 27.00 & 21.75 & 30.38 & \textcolor{blue}{31.83} & \textcolor{blue}{31.53} & \underline{32.78} & \textbf{33.12} \\
SSIM↑ & 0.639 & 0.646 & 0.623 & 0.850 & 0.696 & 0.904 & \textcolor{blue}{0.921} & \textcolor{blue}{0.921} & \underline{0.930} & \textbf{0.933} \\
\end{tabular}}
\\[0.5ex]

\resizebox{\textwidth}{!}{
\begin{tabular}{l|ccccccccccccc}
\hline
\multicolumn{14}{c}{\textbf{The NHR dataset}~\cite{zhang2020nighttime}} \\
\hline
&NDIM$^\dagger$ \cite{zhang2014nighttime}
&GS$^\dagger$ \cite{li2015nighttime}
&MRPF$^\dagger$ \cite{zhang2017fast}
&MRP$^\dagger$ \cite{zhang2017fast}
&OSFD$^\dagger$ \cite{zhang2020nighttime}
&HCD \cite{wang2024restoring}
&FocalNet \cite{cui2023focal}
&Jin \textit{et al}$^\dagger$ \cite{jin2023enhancing}
&\textcolor{blue}{FSNet} \cite{cui2023image}
&\textcolor{blue}{ConvIR-B} \cite{cui2024revitalizing}
&PoolNet-B \cite{cui2025exploring}
&\textbf{FSNet + UDP}
&\textbf{ConvIR + UDP} \\
\midrule
PSNR↑ 
& 14.31 & 17.32 & 16.95 & 19.93 & 21.32 & 23.43 & 25.35 & 26.56 
& \textcolor{blue}{26.30} & \textcolor{blue}{\underline{29.49}} 
& 28.28 & 28.09 & \textbf{29.54} \\
SSIM↑ 
& 0.526 & 0.629 & 0.667 & 0.777 & 0.804 & 0.953 & 0.969 & 0.890 
& \textcolor{blue}{0.976} & \textcolor{blue}{\underline{0.983}} 
& 0.980 & 0.980 & \textbf{0.983} \\
\bottomrule
\end{tabular}}
\\[0.5ex]
\parbox[t]{\textwidth}{\footnotesize
$^\dagger$ Denotes methods that are specially designed for nighttime image dehazing.}
\label{tab:gta5nhr}
\end{table*}

\begin{figure}[t]
    \centering
    \includegraphics[width=1\linewidth]{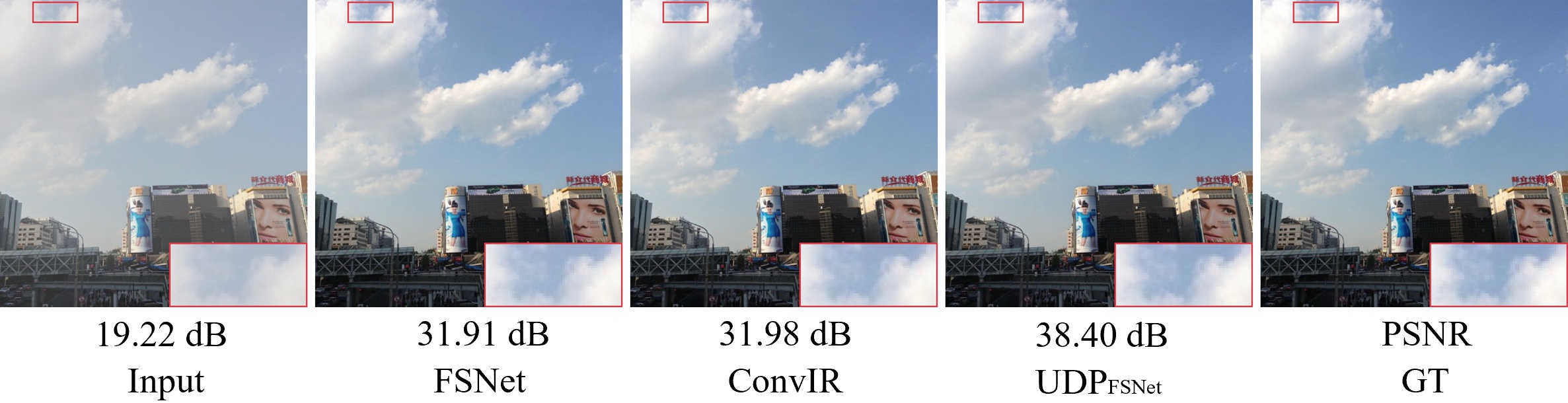}
    \caption{Image dehazing comparisons on the Haze4K dataset~\cite{liu2021synthetic}.}
    \label{fig:haze4k}
\end{figure}

\begin{figure}[t]
    \centering
    \includegraphics[width=1\linewidth]{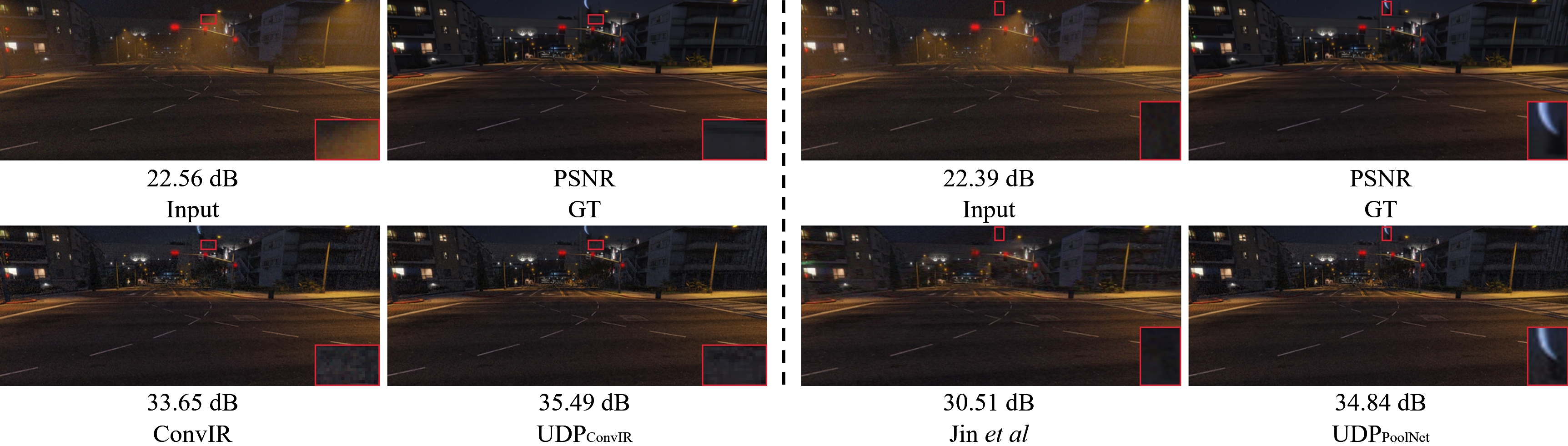}
    \caption{Nighttime image dehazing comparisons on the GTA5 dataset~\cite{yan2020nighttime}.}
    \label{fig:gta5}
\end{figure}

\begin{figure*}[t]
    \centering
    \includegraphics[width=1\linewidth]{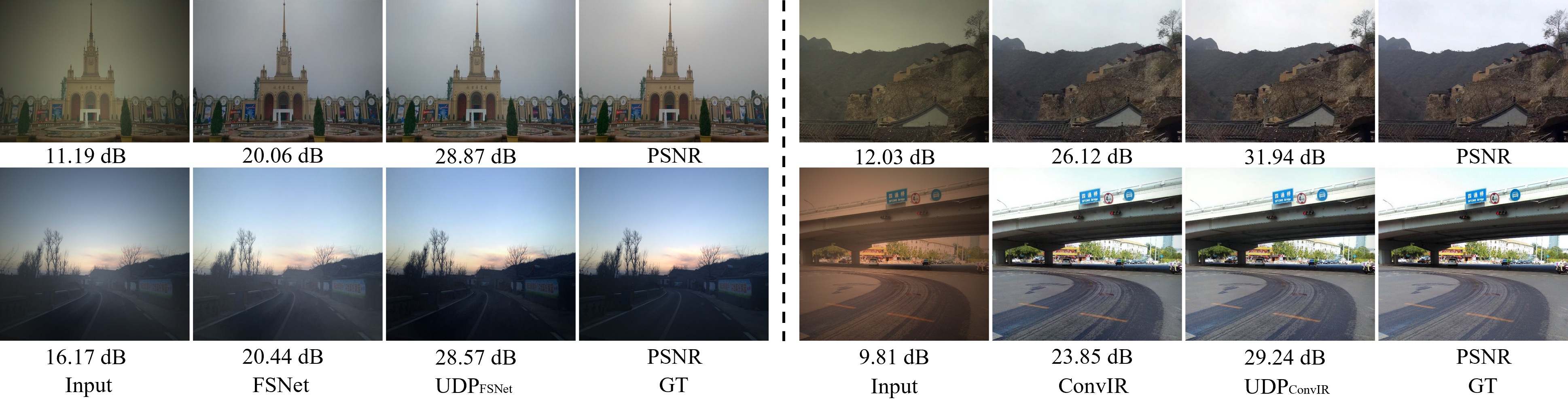}
    \caption{Nighttime image dehazing comparisons on the NHR dataset~\cite{zhang2020nighttime}.}
    \label{fig:nhr}
\end{figure*}

\begin{table}[t]
\centering
\caption{Quantitative results on real-world datasets.}
\resizebox{\columnwidth}{!}{
\begin{tabular}{l|ccc|ccc}
\toprule
\multirow{2}{*}{Method} & \multicolumn{3}{c|}{Dense-Haze~\cite{ancuti2019dense}} & \multicolumn{3}{c}{NH-HAZE~\cite{ancuti2020nh}} \\
 & PSNR↑ & SSIM↑ & LPIPS↓ & PSNR↑ & SSIM↑ & LPIPS↓ \\
\midrule
DehazeNet~\cite{cai2016dehazenet}       & 13.84 & 0.43 & - & 16.62 & 0.52 & - \\
MSBDN~\cite{dong2020multi}              & 15.37 & 0.49 & - & 19.23 & 0.71 & - \\
DeHamer~\cite{guo2022image}             & 16.62 & 0.56 & 0.6346 & 20.66 & 0.68 & 0.3837 \\
PMNet~\cite{ye2022perceiving}           & 16.79 & 0.51 & - & 20.42 & 0.73 & - \\
MB-TaylorFormer-B~\cite{qiu2023mb}      & 16.66 & 0.56 & 0.6125 & -     & -    & - \\
C\textsuperscript{2}PNet~\cite{zheng2023curricular} & 16.88 & 0.57 & - & 20.24 & 0.69 & - \\
FocalNet~\cite{cui2023focal}            & 17.07 & 0.63 & 0.6087 & 20.43 & 0.79 & 0.3780 \\
SFNet~\cite{cui2023selective}           & 17.46 & 0.58 & \textbf{0.5689} & 20.46 & 0.80 & -  \\
\textcolor{blue}{FSNet}~\cite{cui2023image}   & \textcolor{blue}{17.13} & \textcolor{blue}{0.65} & \textcolor{blue}{\underline{0.5756}} & \textcolor{blue}{20.55} & \textcolor{blue}{0.81} & \underline{\textcolor{blue}{0.3624}} \\
\textcolor{blue}{ConvIR-S}~\cite{cui2024revitalizing}     & \textcolor{blue}{17.45} & \textcolor{blue}{0.65} & \textcolor{blue}{0.6000} & \textcolor{blue}{20.65} & \textcolor{blue}{0.80} & \textcolor{blue}{0.3669} \\
PGH$^2$Net~\cite{su2025prior}           & 17.02 & 0.61 & - & - & - & - \\
MB-TaylorFormerV2-B~\cite{jin2025mb}    & 16.95 & 0.62 & - & 20.73 & 0.70 & - \\
\midrule
\textbf{ConvIR + UDP (Ours)}            & \underline{17.55} & \textbf{0.67} & 0.5813 & \textbf{20.98} & \textbf{0.82} & \textbf{0.3567} \\
\textbf{FSNet + UDP (Ours)}             & \textbf{17.85} & \underline{0.65} & 0.6033 & \underline{20.94} & \underline{0.82} & 0.3732 \\
\bottomrule
\end{tabular}
}
\label{tab:real-world}
\end{table}

\begin{figure*}[t]
    \centering
    \includegraphics[width=1\linewidth]{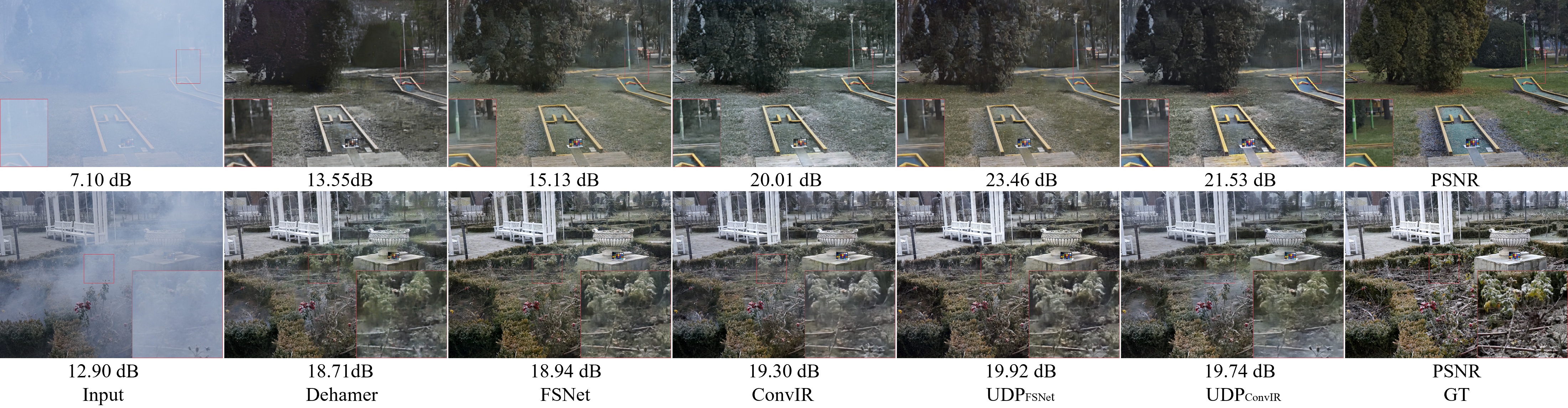}
    \caption{Qualitative comparisons on the Dense-Haze dataset~\cite{ancuti2019dense} and NH-HAZE dataset~\cite{ancuti2020nh} among DeHamer~\cite{guo2022image}, FSNet~\cite{cui2023image}, ConvIR~\cite{cui2024revitalizing}, and ours.}
    \label{fig:real-world}
\end{figure*}

\begin{table}[t]
\centering
\caption{Image Dehazing Comparisons on the Remote Sensing Datasets~\cite{huang2020single}}
\resizebox{\columnwidth}{!}{
\begin{tabular}{l|cc|cc|cc}
\toprule
\multirow{2}{*}{Method} 
  & \multicolumn{2}{c|}{Thin} 
  & \multicolumn{2}{c|}{Moderate} 
  & \multicolumn{2}{c}{Thick} \\
  & PSNR↑ & SSIM↑ & PSNR↑ & SSIM↑ & PSNR↑ & SSIM↑ \\
\midrule
AOD-Net~\cite{li2017aod}                            & 19.54 & 0.854 & 20.10 & 0.885 & 15.92 & 0.731 \\
H2RL-Net$^\dagger$~\cite{chen2021hybrid}            & 20.91 & 0.880 & 22.34 & 0.906 & 17.41 & 0.768 \\
FCFT-Net$^\dagger$~\cite{li2020coarse}              & 23.59 & 0.913 & 22.88 & 0.927 & 20.03 & 0.816 \\
C\textsuperscript{2}PNet~\cite{zheng2023curricular} & 19.62 & 0.880 & 24.79 & 0.940 & 16.83 & 0.790 \\
Restormer~\cite{zamir2022restormer}                 & 23.08 & 0.912 & 24.73 & 0.933 & 18.58 & 0.762 \\
Trinity-Net$^\dagger$~\cite{chi2023trinity}         & 21.55 & 0.884 & 23.35 & 0.895 & 20.97 & 0.823 \\
UMWTransformer~\cite{kulkarni2022unified}           & 24.29 & 0.919 & 26.65 & 0.946 & 20.07 & 0.825 \\
FocalNet~\cite{cui2023focal}                        & 24.16 & 0.916 & 25.99 & 0.947 & 21.69 & 0.847 \\
\textcolor{blue}{ConvIR-S}~\cite{cui2024revitalizing} & \textcolor{blue}{25.11} & \textcolor{blue}{0.978} & \textcolor{blue}{26.79} & \textcolor{blue}{0.978} & \textcolor{blue}{22.65} & \textcolor{blue}{0.950} \\
\textcolor{blue}{PoolNet-S}~\cite{cui2025exploring}   & \textcolor{blue}{25.02} & \textcolor{blue}{0.979} & \textcolor{blue}{27.02} & \textcolor{blue}{0.979} & \textcolor{blue}{22.73} & \textcolor{blue}{\textbf{0.955}} \\
\midrule
\textbf{ConvIR + UDP (Ours)}       & \underline{25.48} & \underline{0.979} & \underline{28.07} & \textbf{0.981} & \underline{22.95} & \underline{0.953} \\
\textbf{PoolNet + UDP (Ours)}      & \textbf{26.20} & \textbf{0.980} & \textbf{28.26} & \underline{0.979} & \textbf{23.13} & 0.951 \\
\bottomrule
\end{tabular} 
}
\label{tab:remote-sensing}
\\[0.5ex]
\parbox[t]{\textwidth}{\footnotesize
$^\dagger$ Denotes methods that are designed for remote sensing image dehazing.}
\end{table}

\subsection{Comparisons with SOTA Methods}
We first present a comprehensive comparison with recent approaches on the SOTS dataset~\cite{li2018benchmarking} in Table~\ref{tab:dehaze_all}. As shown, our proposed methods consistently achieve superior performance across both datasets. In particular, ConvIR + UDP obtains a remarkable PSNR of 43.12 dB on SOTS-Indoor, surpassing the previous best ConvIR-B~\cite{cui2024revitalizing} and MB-TaylorFormerV2-L~\cite{jin2025mb} by 0.40 dB and 0.28 dB, respectively. On the SOTS-Outdoor dataset, it also achieves a competitive 40.32 dB PSNR, outperforming most recent models such as C\textsuperscript{2}PNet~\cite{zheng2023curricular}, and DEA-Net~\cite{chen2024dea}. More importantly, our FSNet + UDP further pushes the performance boundary, setting a new state-of-the-art with 43.30 dB PSNR on SOTS-Indoor and 40.53 dB PSNR on SOTS-Outdoor. These results highlight the effectiveness of our UDP design, achieving a new milestone in image dehazing. Moreover, we further evaluate our approach on the more realistic synthetic Haze4K dataset~\cite{liu2021synthetic}. Table~\ref{tab:haze4k} showcases the performance of our UDP method applied to FSNet. As reported in Table~\ref{tab:haze4k}, applying our UDP to FSNet yields an average PSNR improvement of 1.19 dB over the original model. In addition, ConvIR + UDP also achieves a competitive PSNR of 34.82 dB. The qualitative comparisons on these daytime datasets, are illustrated in Figs.~\ref{fig:iots} and \ref{fig:haze4k}, respectively. Our models generate haze-free images that are more visually consistent with the GT.

To further demonstrate the superiority of our model, we conduct experiments on two nighttime datasets, \ieno, GTA5~\cite{yan2020nighttime} and NHR~\cite{zhang2020nighttime}. Table~\ref{tab:gta5nhr} shows that ConvIR + UDP and PoolNet + UDP secure the first and second places with 33.12 dB and 32.78 dB PSNR, surpassing the original counterparts by 1.29 dB and 1.25 dB, respectively. The original models struggle to achieve optimal results under low-light conditions. However, by incorporating our UDP approach, we effectively alleviate these limitations, enhancing the overall brightness and recovering fine image details. The visual comparisons in Fig.~\ref{fig:gta5} further confirm the effectiveness of our design, as the images generated by our models are much closer to the ground truth. We further extensively compare our models with state-of-the-art methods on real-world datasets. As shown in Table~\ref{tab:real-world}, our methods achieve state-of-the-art performance on both Dense-Haze~\cite{ancuti2019dense} and NH-HAZE~\cite{ancuti2020nh}. On Dense-Haze, FSNet + UDP obtains 17.85 dB PSNR, surpassing FSNet by 0.72 dB. On NH-HAZE, ConvIR + UDP achieves the best performance with 20.98 dB PSNR, outperforming the previous best MB-TaylorFormerV2-B~\cite{jin2025mb}, and also attains the best perceptual quality. These results clearly demonstrate the effectiveness of our UDP in boosting real-world dehazing performance. The visual comparisons are shown in Fig.~\ref{fig:real-world}.

Image dehazing is crucial for remote sensing applications, we further evaluate our models on SateHaze1k~\cite{huang2020single} in Table~\ref{tab:remote-sensing}. Our methods consistently achieve the best results across all subsets. In particular, PoolNet + UDP attains 26.20 dB, 28.26 dB, and 23.13 dB PSNR on the three subsets, outperforming both general dehazing models (\egno, FocalNet~\cite{cui2023focal}, ConvIR-S~\cite{cui2024omni}) and remote sensing–oriented designs (\egno, Trinity-Net~\cite{chi2023trinity}). Notably, on the challenging Thick subset, our method achieves a gain of 1.44 dB over FocalNet, highlighting its strong generalization for remote sensing scenarios.

To further demonstrate the superiority of our approach, we conduct experiments on the all-in-one settings~\cite{jiang2024survey}. The settings evaluate the ability of a single model to function as a universal and versatile restorer. In our study, we adopt the more challenging five-task benchmark, which jointly considers Gaussian denoising, deraining, dehazing, motion deblurring, and low-light enhancement, thus requiring stronger generalization from a single set of model weights. As shown in Table~\ref{tab:allinone}, both PromptIR + UDP and AdaIR + UDP achieve substantial improvements in dehazing and low-light enhancement, with PromptIR + UDP yields further gains in deraining and AdaIR + UDP further boosting performance in motion deblurring. This can be largely attributed to the integration of depth-based priors, which promote a more nuanced understanding of global atmospheric illumination and fine-grained object structures. Although originally introduced to enhance dehazing, such priors also strengthen the model’s ability to recover structural fidelity in motion deblurring and illumination details in low-light enhancement. Consequently, AdaIR + UDP surpasses its original counterpart and establishes a new benchmark for all-in-one image restoration.

\begin{table*}[htp]
\centering
\caption{Performance comparison of different methods on five challenging benchmarks. Denoising results are reported at $\sigma=25$. The baseline results are in blue. Best results are \textbf{highlighted}.}
\label{tab:allinone}
\resizebox{\textwidth}{!}{
\begin{tabular}{l|cc|cc|cc|cc|cc|cc}
\Xhline{1.1pt}
\multirow{2}{*}{Method} & \multicolumn{2}{c|}{Dehazing on SOTS\cite{li2018benchmarking}} & \multicolumn{2}{c|}{Deraining on Rain100L\cite{yang2019joint}} & \multicolumn{2}{c|}{Denoising on BSD68\cite{martin2001database}} & \multicolumn{2}{c|}{Deblurring on GoPro\cite{nah2017deep}} & \multicolumn{2}{c|}{Low-Light on LOL\cite{wei2018deep}} & \multicolumn{2}{c}{Average} \\
 & PSNR & SSIM & PSNR & SSIM & PSNR & SSIM & PSNR & SSIM & PSNR & SSIM & PSNR & SSIM \\
\Xhline{1.1pt}
(TIP'23) DehazeFormer~\cite{song2023vision} & 25.31 & 0.937 & 33.68 & 0.954 & 30.89 & 0.880 & 25.93 & 0.785 & 21.31 & 0.819 & 27.42 & 0.875 \\
(ICCV'23) Retinexformer~\cite{cai2023retinexformer} & 24.81 & 0.933 & 32.68 & 0.940 & 30.84 & 0.880 & 25.09 & 0.779 & 22.76 & 0.863 & 27.24 & 0.873 \\
\hline
(ICCVW’21) SwinIR~\cite{liang2021swinir} & 21.50 & 0.891 & 30.78 & 0.923 & 30.59 & 0.868 & 24.52 & 0.773 & 17.81 & 0.723 & 25.04 & 0.835 \\
(CVPR'22) Restormer~\cite{zamir2022restormer} & 24.09 & 0.927 & 34.81 & 0.960 & 31.49 & 0.884 & 27.22 & 0.829 & 20.41 & 0.806 & 27.60 & 0.881 \\
(TPAMI'23) FSNet~\cite{cui2023image} & 25.53 & 0.943 & 36.07 & 0.968 & 31.33 & 0.883 & 28.32 & 0.869 & 22.29 & 0.829 & 28.71 & 0.898 \\
\hline
(CVPR'22) AirNet~\cite{li2022all} & 21.04 & 0.884 & 32.98 & 0.951 & 30.91 & 0.882 & 24.35 & 0.781 & 18.18 & 0.735 & 25.49 & 0.846 \\
\textcolor{blue}{(NeurIPS'23) PromptIR}~\cite{potlapalli2023promptir} & \textcolor{blue}{26.54} & \textcolor{blue}{0.949} & \textcolor{blue}{36.37} & \textcolor{blue}{0.970} & \textcolor{blue}{\textbf{31.47}} & \textcolor{blue}{0.886} & \textcolor{blue}{28.71} & \textcolor{blue}{0.881} & \textcolor{blue}{22.68} & \textcolor{blue}{0.832} & \textcolor{blue}{29.15} & \textcolor{blue}{0.904} \\
\textcolor{blue}{(ICLR'25) AdaIR}~\cite{cui2025adair} & \textcolor{blue}{30.53} & \textcolor{blue}{0.978} & \textcolor{blue}{\underline{38.02}} & \textcolor{blue}{\underline{0.981}} & \textcolor{blue}{31.35} & \textcolor{blue}{\underline{0.889}} & \textcolor{blue}{28.12} & \textcolor{blue}{0.858} & \textcolor{blue}{23.00} & \textcolor{blue}{0.845} & \textcolor{blue}{30.20} & \textcolor{blue}{0.910} \\
(ICLR'25) $\text{DCPT}_\text{PromptIR}$~\cite{hu2025universal} & 30.72 & 0.977 & 37.32 & 0.978 & 31.32 & 0.885 & \underline{28.84} & \underline{0.877} & \underline{23.35} & 0.840 & 30.31 & 0.911 \\
(TPAMI'25) DA-RCOT~\cite{tang2025degradation} & 30.96 & 0.975 & 37.87 & 0.980 & 31.23 & 0.888 & 28.68 & 0.872 & 23.25 & 0.836 & \underline{30.40} & 0.911 \\
(TIP'25) Perceive-IR~\cite{zhang2025perceive} & 28.19 & 0.964 & 37.25 & 0.977 & \underline{31.44} & 0.887 & \textbf{29.46} & \textbf{0.886} & 22.88 & 0.833 & 29.84 & 0.909 \\
(TIP'25) Pool-AIO~\cite{cui2025exploring} & 30.25 & 0.977 & 37.85 & 0.981 & 31.24 & 0.887 & 27.66 & 0.844 & 22.66 & 0.841 & 29.93 & 0.906 \\
(TIP'25) $\text{DPPD}_\text{PromptIR}$~\cite{wu2025learning} & 30.31 & 0.980 & 37.32 & 0.980 & 31.33 & 0.885 & 28.74 & 0.875 & 22.73 & 0.846 & 30.09 & \underline{0.913} \\
(CVPR'25) VLU-Net~\cite{zeng2025vision} & 30.84 & 0.980 & \textbf{38.54} & \textbf{0.982} & 31.43 & \textbf{0.891} & 27.46 & 0.840 & 22.29 & 0.833 & 30.11 & 0.905 \\
\textbf{(Ours) PromptIR + UDP} & \underline{31.33} & \underline{0.980} & 37.63 & 0.980 & 31.25 & 0.883 & 28.34 & 0.868 & 23.18 & \underline{0.851} & 30.35 & 0.912\\
\textbf{(Ours) AdaIR + UDP} & \textbf{31.41} & \textbf{0.980} & 37.85 & 0.980 & 31.28 & 0.888 & 28.62 & 0.870 & \textbf{23.53} & \textbf{0.854} & \textbf{30.55} & \textbf{0.915} \\
\Xhline{1.1pt}
\end{tabular}}
\end{table*}

\begin{table*}[t]
	\centering
	\caption{Ablation analysis on different variants of our method, mainly including two aspects: fusion stages and attention types. Here, all ablation models adopt the backbone of FSNet~\cite{cui2023image}.}
	\vspace{-2mm}
	\resizebox{1.0\textwidth}{!}{
	\begin{tabular}{cccccccccccccc}
	\toprule
	\multirow{2}{*}{} 
	  & \multicolumn{2}{c}{Head} 
	  & \multicolumn{4}{c}{Encoder}          
	  & \multicolumn{4}{c}{Decoder}       
	  & \multicolumn{1}{c}{Backbone}
	  & \multicolumn{2}{c}{Metrics} \\ 
	\cmidrule(lr){2-3} \cmidrule(lr){4-7} \cmidrule(lr){8-11} \cmidrule(lr){12-12} \cmidrule(lr){13-14}
	  & concat     & DGAM    
	  & \text{CCA\textsubscript{depth}}  & \text{CCA\textsubscript{rgb}} & \text{SCA\textsubscript{depth}} & \text{SCA\textsubscript{rgb}} 
	  & \text{CCA\textsubscript{depth}}  & \text{CCA\textsubscript{rgb}} & \text{SCA\textsubscript{depth}} & \text{SCA\textsubscript{rgb}} 
	  & FSNet 
	  & PSNR     & SSIM      \\ 
	\midrule
	(a)                        
	  & \xmark              & \xmark             
	  & \xmark & \xmark & \xmark & \xmark            
	  & \xmark & \xmark & \xmark & \xmark               
	  & \cmark                         
	  & 34.12  & 0.9901    \\
	(b)                        
	  & \cmark              & \xmark             
	  & \xmark & \xmark & \xmark & \xmark            
	  & \xmark & \xmark & \xmark & \xmark               
	  & \cmark                         
	  & 34.78    & 0.9908    \\
	(c)                        
	  & \xmark              & \cmark             
	  & \xmark & \xmark & \xmark & \xmark            
	  & \xmark & \xmark & \xmark & \xmark               
	  & \cmark                         
	  & 34.96    & 0.9908  \\
	(d)                        
	  & \xmark              & \cmark             
	  & \cmark & \xmark & \xmark & \xmark            
	  & \xmark & \xmark & \xmark & \xmark               
	  & \cmark                         
	  & 34.93   & 0.9908    \\
	(e)                        
	  & \xmark              & \cmark             
	  & \xmark & \cmark & \xmark & \xmark            
	  & \xmark & \xmark & \xmark & \xmark               
	  & \cmark                         
	  & 35.06    & 0.9910    \\
	(f)                        
	  & \xmark              & \cmark             
	  & \xmark & \xmark & \cmark & \xmark            
	  & \xmark & \xmark & \xmark & \xmark               
	  & \cmark                         
	  & 35.10    & 0.9907    \\
	(g)                        
	  & \xmark              & \cmark             
	  & \xmark & \xmark & \xmark & \cmark            
	  & \xmark & \xmark & \xmark & \xmark               
	  & \cmark                         
	  & 35.13    & 0.9910    \\ 
        (h)                        
	  & \xmark              & \cmark             
	  & \xmark & \xmark & \xmark & \xmark            
	  & \cmark & \xmark & \xmark & \xmark               
	  & \cmark                         
	  & 34.94   & 0.9909    \\
	(i)                        
	  & \xmark              & \cmark             
	  & \xmark & \xmark & \xmark & \xmark            
	  & \xmark & \cmark & \xmark & \xmark               
	  & \cmark                         
	  & 34.99   & 0.9907    \\
	(j)                        
	  & \xmark              & \cmark             
	  & \xmark & \xmark & \xmark & \xmark            
	  & \xmark & \xmark & \cmark & \xmark               
	  & \cmark                         
	  & 34.92    & 0.9909    \\
	(k)                        
	  & \xmark              & \cmark             
	  & \xmark & \xmark & \xmark & \xmark            
	  & \xmark & \xmark & \xmark & \cmark               
	  & \cmark                         
	  & 35.02    & 0.9910    \\ 
        (l)                        
	  & \xmark              & \cmark             
	  & \xmark & \xmark & \cmark & \cmark            
	  & \xmark & \xmark & \xmark & \xmark               
	  & \cmark                         
	  & 35.22    & 0.9908    \\ 
	\bottomrule
	\end{tabular}}
	\vspace{-4mm}
	\label{tab:ablation}
\end{table*}

\subsection{Ablation Study}
\label{sec:Ablation_Study}
We conduct ablation studies to demonstrate the effectiveness of the proposed UDPNet by training and testing the models on the Haze4K dataset~\cite{liu2021synthetic}. The model is trained for 1000 epochs using FSNet~\cite{cui2023image} as the baseline.

\subsubsection{Contribution of the Depth Map}
The first step in our analysis is to justify the integration of depth information. 
Table~\ref{tab:depth_map} shows that the baseline FSNet achieves 34.12 dB PSNR. By incorporating depth maps extracted using the DepthAnything-V2-small variant together with our proposed UDP, the model’s performance is improved to 34.69 dB, while only introducing 0.32 M parameters. Although the gain at this stage is relatively modest, it demonstrates the effectiveness of depth priors. 
Using the base and large variants of DepthAnything V2 further increases performance, reaching 34.87 dB and a peak of 35.31 dB, respectively. In contrast, when the depth map is replaced with a 128-level gray map, the performance drops to 34.08 dB, highlighting that the performance improvement comes from accurate depth information rather than the increase in parameter count. Overall, these results confirm that accurate depth estimation is a key factor in image dehazing.

\begin{table}[t]
\centering
\caption{Comparison of different depth maps. All depth variants have an identical parameter increase of only +0.32M.}
\label{tab:depth_map}
\resizebox{\columnwidth}{!}{
\begin{tabular}{lccccc}
\toprule
Methods & Baseline (13.28M) & Grey (13.60M) & Small (13.60M) & Base (13.60M) & Large (13.60M) \\
\midrule
PSNR & 34.12 dB & 34.08 dB  & 34.69 dB  & 34.87 dB  & 35.31 dB  \\
\bottomrule
\end{tabular}
}
\end{table}

\begin{figure}[htbp]
    \centering
    \includegraphics[width=1\linewidth]{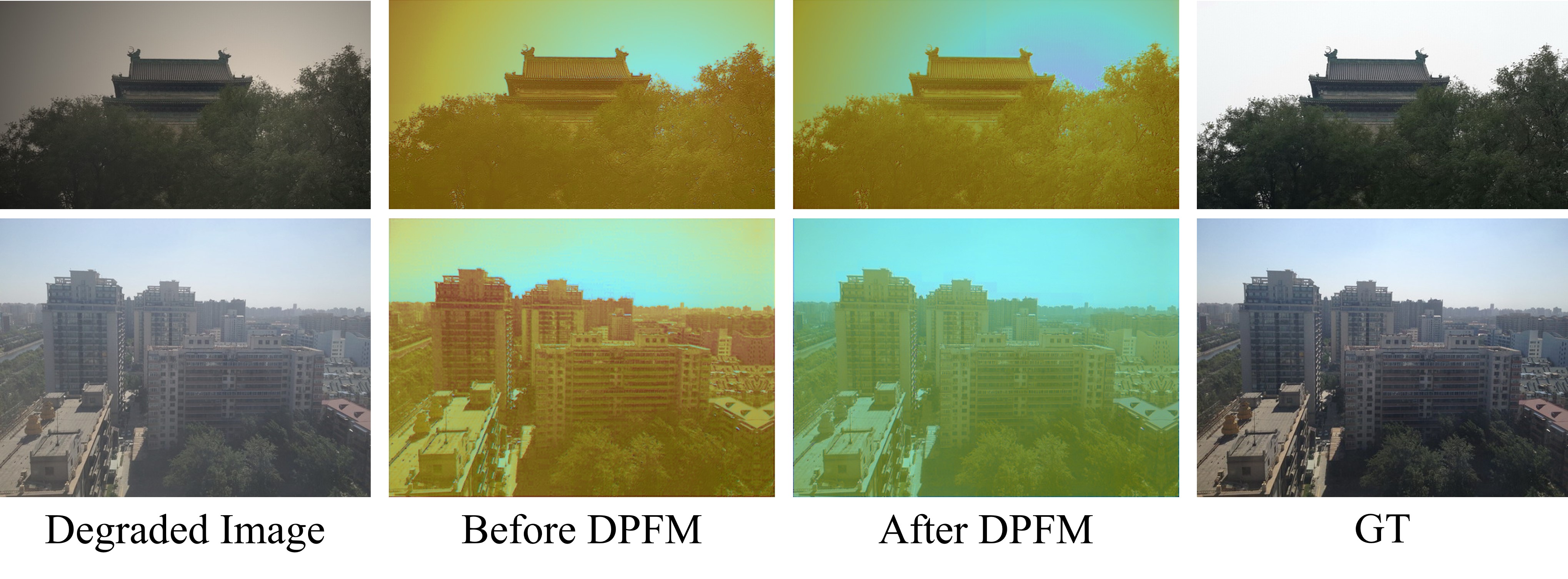}
    \caption{Illustrations of intermediate features. The proposed DPFM effectively generates features with enhanced textures, sharper edges, and clearer global structures, thereby facilitating high-quality reconstruction.}
    \label{fig:vis}
\end{figure}

\begin{figure}[t]
    \centering
    \includegraphics[width=1\linewidth]{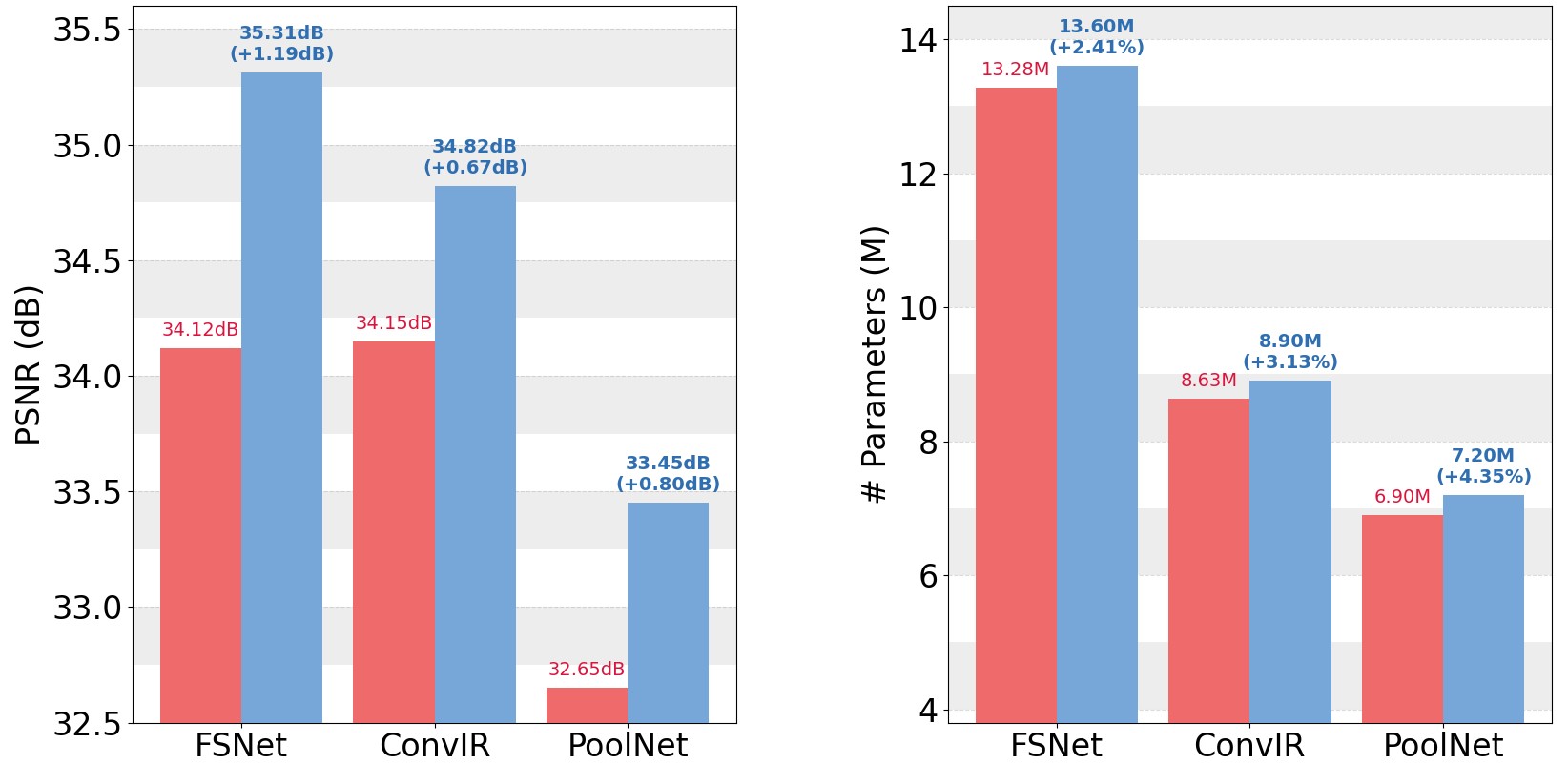}
    \caption{Performance and efficiency variations with our UDP on the Haze4K dataset, compared with baseline methods~\cite{cui2023image,cui2024revitalizing,cui2025exploring}.}
    \label{fig:cmp}
\end{figure}

\subsubsection{Effectiveness of each module}
We perform the breakdown ablations by applying our modules to the baseline successively. The number of heads in the multi-head attention is set to 1, while other configurations remain identical to the final dehazing model. As shown in Table~\ref{tab:ablation}, directly concatenating refined depth maps and RGB images at the network head (row (b)) significantly improves performance compared to the baseline (row (a)), reaching 34.78 dB PSNR. This demonstrates the substantial benefit of introducing depth maps. By contrast, replacing simple concatenation with the proposed DGAM (row (c)) yields further gains with 34.96 dB PSNR, verifying that adaptively weighting depth cues through channel attention is more effective than naive feature stacking.  

We further explore two variants of DPFM: channel cross-attention (CCA) and spatial cross-attention (SCA). Results in rows (d–g) show that both variants outperform DGAM, indicating the benefit of explicitly modeling cross-modal dependencies. Notably, using RGB features as the query (CCA\textsubscript{rgb}, SCA\textsubscript{rgb}) consistently outperforms the depth-as-query counterparts, suggesting that RGB semantics provide a more reliable guidance signal. Moreover, spatial cross-attention (SCA) shows slightly higher gains than channel cross-attention (CCA), suggesting that spatial correspondences play a crucial role in depth–RGB fusion. Across all cases, the best performance 35.22 dB PSNR is achieved in row (l), where both SCA\textsubscript{depth} and SCA\textsubscript{rgb} are combined in the encoder, highlighting the complementarity of depth and RGB at the spatial level and confirming the necessity of deliberate design.

Finally, we provide a visualization of the intermediate features to further demonstrate the effectiveness of the proposed DPFM, as shown in Figure~\ref{fig:vis}. Specifically, the features are extracted from the first encoder before and after applying our DPFM. Moreover, Figure~\ref{fig:cmp} illustrates that our UDP, as a plugin, can be seamlessly integrated into off-the-shelf methods to improve their performance while avoiding a significant increase in the number of parameters.

\subsubsection{Position of DPFM}
We also investigate inserting DPFM into different network stages. Placing DPFM in the encoder (rows (d–g)) yields more notable gains in Table~\ref{tab:ablation}, since incorporating depth priors at early stages enhances semantic abstraction and provides stronger guidance for subsequent layers. By contrast, applying DPFM to the decoder (rows (h–k)) still brings consistent improvements, but the benefits are relatively smaller. This can be attributed to the fact that depth maps mainly contain low-frequency structural cues, which may interfere with the reconstruction of high-frequency image details when introduced in the decoder. 

\subsubsection{Impact of the number of heads in DPFM}
We investigate the effect of varying the number of attention heads in the multi-head overlapping cross-attention of DPFM. Specifically, we evaluate configurations with 1, 2, and 4 heads. When using a single head, the model achieves 35.20 dB PSNR. Increasing the number of heads to 2 yields the best performance of 35.31 dB PSNR, suggesting that multiple attention heads can capture complementary correlations between RGB and depth features more effectively. Further increasing to 4 heads slightly decreases the PSNR to 35.27 dB. Overall, these results demonstrate that a moderate number of attention heads is sufficient to fully exploit cross-modal dependencies while maintaining high-frequency detail reconstruction.

\subsubsection{Generalization of our proposed UDP}
To evaluate the adaptability of the proposed UDP, we integrate it into different baseline models, including FSNet~\cite{cui2023image}, ConvIR~\cite{cui2024revitalizing}, PoolNet~\cite{cui2025exploring}, PromptIR~\cite{potlapalli2023promptir} and AdaIR~\cite{cui2025adair}. Each modified network is trained and tested under the corresponding experimental settings. The results demonstrate that UDP can be applied to a variety of multi-scale image restoration networks, while effectively enhance the performance of all-in-one image restoration models, yielding improvements in tasks such as dehazing, deblurring, deraining, and low-light enhancement.

\subsection{Limitations}
Our work introduces an effective method for image dehazing by leveraging depth-based priors, which significantly enhance adaptability and performance through plug-and-play modules. By integrating depth features, our approach lays the foundation for developing more robust and versatile dehazing models. However, the UDP has several limitations that open avenues for future research. One limitation is the reliance on depth priors extracted from degraded images, which may introduce inaccuracies due to noise or occlusions in the depth estimation process. These inaccuracies could lead to structural distortions or loss of fine details in the restored image. Furthermore, the dependence on DepthAnything V2 priors during inference introduces computational inefficiency, as the external model requires additional resources for depth estimation, increasing inference time. Future work could explore alternative strategies for real-time depth inference or distilling depth priors from depth estimation models into image restoration models. While the UDP framework demonstrates promising results across various restoration tasks, the potential of leveraging depth-based priors for general image restoration remains an exciting direction. Future studies will focus on refining adaptive strategies to improve performance in a broader range of scenarios, with particular attention to enhancing real-time performance.

\section{Conclusion}
\label{sec:conclusion}
In recent years, the field of single-image dehazing has witnessed significant advancements, primarily driven by effective module design. However, the utilization of multimodal information remains limited, and depth information that is closely tied to the physical process of haze formation is often overlooked. In this paper, we present UDPNet, a general framework guided by depth priors for image dehazing to enhance multi-scale and dual-modal representation learning. Our approach integrates a large-scale pre-trained depth estimation model to extract scene depth priors, which are utilized within diverse U-shaped network architectures. By leveraging efficient fusion strategies, we adaptively tailor the model to handle different haze distributions. These combined innovations yield a robust solution for image dehazing, as demonstrated by our extensive evaluations, while also showcasing the potential of our framework for universal image restoration.

\bibliographystyle{IEEEtran}
\bibliography{ref}

\vfill

\end{document}